\documentclass{article}
\usepackage{hyperref}

\usepackage{./statistics-macros}
\usepackage{pgfplotstable}
\usepackage{float}

\usepackage{microtype}
\usepackage{subfigure}
\usepackage{booktabs} 

\usepackage[arxiv]{icml2019}

\usepackage{overpic}
\usepackage{tikz}
\usepackage{rotating}
\usepackage{psfrag}
\usepackage{xcolor,cancel}

\definecolor{innerboxcolor}{rgb}{.9,.95,1}
\definecolor{outerlinecolor}{rgb}{.6,0,.2}

\providecommand{\comment}[1]{}

\newcommand{\outcomes}{W}

\makeatletter
\long\def\@makecaption#1#2{
  \vskip 0.8ex
  \setbox\@tempboxa\hbox{\small {\bf #1:} #2}
  \parindent 1.5em  
  \dimen0=\hsize
  \advance\dimen0 by -3em
  \ifdim \wd\@tempboxa >\dimen0
  \hbox to \hsize{
    \parindent 0em
    \hfil 
    \parbox{\dimen0}{\def\baselinestretch{0.96}\small
      {\bf #1.} #2
    } 
    \hfil}
  \else \hbox to \hsize{\hfil \box\@tempboxa \hfil}
  \fi
}
\makeatother

\icmltitlerunning{Off-policy Policy Evaluation Under Unobserved Confounding}

\begin{document}

\twocolumn[ \icmltitle{Off-policy Policy Evaluation For Sequential Decisions \\
  Under Unobserved Confounding}



\icmlsetsymbol{equal}{*}

\begin{icmlauthorlist}
  \icmlauthor{Hongseok Namkoong}{equal} 
  \icmlauthor{Ramtin Keramati}{equal,rest}
  \icmlauthor{Steve Yadlowsky}{equal,rest} 
\icmlauthor{Emma Brunskill}{rest}
\end{icmlauthorlist}

\icmlaffiliation{rest}{Stanford University, Stanford, CA, USA}

\icmlcorrespondingauthor{Steve Yadlowsky}{syadlows@stanford.edu}

\icmlkeywords{off-policy learning, policy evaluation, confounding, sensitivity
  analysis}

\vskip 0.3in
]



\printAffiliationsAndNotice{\icmlEqualContribution} 


\begin{abstract}
  When observed decisions depend only on observed features, off-policy policy
evaluation (OPE) methods for sequential decision making problems can estimate
the performance of evaluation policies before deploying them. This assumption
is frequently violated due to unobserved confounders, unrecorded variables
that impact both the decisions and their outcomes. We assess robustness of OPE
methods under unobserved confounding by developing worst-case bounds on the
performance of an evaluation policy. When unobserved confounders can affect
every decision in an episode, we demonstrate that even small amounts of
per-decision confounding can heavily bias OPE methods. Fortunately, in a
number of important settings found in healthcare, policy-making, operations,
and technology, unobserved confounders may primarily affect only one of the
many decisions made. Under this less pessimistic model of one-decision
confounding, we propose an efficient loss-minimization-based procedure for
computing worst-case bounds, and prove its statistical consistency. On two
simulated healthcare examples---management of sepsis patients and
developmental interventions for autistic children---where this is a reasonable
model of confounding, we demonstrate that our method invalidates non-robust
results and provides meaningful certificates of robustness, allowing reliable
selection of policies even under unobserved confounding.


\end{abstract}


\newcommand{\disc}{\gamma}

\section{Introduction}
\label{section:introduction}

New technology and regulatory shifts allow the collection of an unprecedented amount
of data on past decisions and their associated outcomes, ranging from product
recommendation systems
to medical treatment decisions. This presents unique opportunities for using
off-policy observational data to inform better decision-making. When online
experimentation is expensive or risky, it is crucial to leverage prior data to
evaluate the performance of a sequential decision policy (which we call the \emph{evaluation policy}) before deploying
it. The dynamic treatment regime literature~\citep{Robins86,
Robins97, Murphy03} addressed many early questions around using observational data for sequential decision making, and developed a rich set of methods adapted for epidemiological questions. The
reinforcement learning (RL) community is increasingly interested in developing theory and methods for the related problem of batch
RL across a broad set of applications, because of new models and data availability (see
e.g.~\citet{thomas2019,liu2018representation,le2019batch,thomas2015high,komorowski2018artificial,hanna2017bootstrapping,gottesman2019guidelines,gottesman2019combining}). We
focus on performing off-policy policy evaluation (OPE) in the common scenario where decisions are made in episodes by an
unknown behavior policy, each involving a sequence of decisions.

A central challenge in OPE is that the estimand is inherently a counterfactual
quantity: what would the resulting
outcomes
be \textit{if an alternate policy
  had been used} (the counterfactual) instead of the behavior policy
that generated the observed data (the factual). As a result, OPE requires
causal reasoning about whether observed high/low rewards were caused by
observed decisions, as opposed to a common causal variable that
simultaneously affects both the observed decisions and the states or
rewards~\cite{HernanRo20,Pearl09}.
In order to make counterfactual evaluations possible, a standard
assumption---albeit often overlooked and unstated---is to require that the
behavior policy does not depend on any unobserved/latent variables that also
affect the future states or rewards (no unobserved confounding). We refer to
this assumption as \emph{sequential ignorability}, following the line of works
on dynamic treatment regimes~\citep{Robins86, Robins97, MurphyVaRo01, Murphy03}.

Sequential ignorability, however, is almost always violated in observational
problems where the behavior policy is unknown. In healthcare, business
operations, and some automated systems in tech, decisions depend on
unlogged features correlated with future outcomes.  Clinicians use visual
observations or discussions with patients to inform treatment decisions, but
such information is typically not quantified and coded in electronic medical
records; they also rely on heuristics that are fundamentally difficult
to quantify, tending to over-extrapolate on past experience and slow to
correct mistakes~\citep{Mcdonald96}. In judicial decisions, psychological factors affect bail and
parole decisions~\citep{Dhami03, DanzigerLeAv11}. In business contexts, simple
heuristics are prevalent; concrete examples include venture capital
investments~\citep{AastebroEl06}, and customer targeting~\citep{WubbenWa08}.
Even automated policies in tech firms depend on unlogged
features~\citep{AgarwalEtAl16}, and complex software and data infrastructures often introduce
 confounding.


 In this paper, we study a framework for quantifying the impact of unobserved
 confounders on OPE estimates, developing worst-case bounds on the performance
 of an evaluation policy.  Since OPE is generally impossible under arbitrary
 unobserved confounding, we begin by positing a model that explicitly limits
 their influence on decisions. Our proposed model is a natural extension of an
 influential confounding model for a single binary decision~\cite{Rosenbaum02}
 to the multi-action sequential decision making setting.  When unobserved
 confounders can affect all decisions, even small amounts of confounding can
 have an exponential (in the number of decisions) impact on the bias of OPE as
 we illustrate in Section~\ref{section:bounds}. In this sense, the validity of
 OPE can almost always be questioned under presence of unobserved confounding
 that affect all decisions.

Fortunately, in a number of important applications, unobserved confounders may
only affect a single decision. Frequently, this happens when a high-level
expert decision-makers make an initial decision potentially using unrecorded
information, after which a standard set of protocols are followed based on
well-recorded observations. Under our less pessimistic model of
single-decision confounding, we develop bounds on the expected cumulative
rewards under the evaluation policy (Section~\ref{section:convex}). We use
functional convex duality to derive a dual relaxation, and show that it can be
computed by solving a loss minimization problem. Our procedure allows
analyzing sensitivity of OPE methods to confounding in realistic scenarios
involving continuous state and rewards, over a potentially large horizon.  We
prove that an empirical approximation of our procedure is consistent, allowing
estimation from observed past decisions. Our loss minimization approach builds
on the single decision work by~\citet{YadlowskyNaBaDuTi18}, and
extends it to sequential decision-making scenarios.

On examples of dynamic treatment regimes for autism and sepsis management, we
illustrate how our single-decision confounding model allows informative bounds
over meaningful amounts of confounding. Our approach provides certificates of
robustness by identifying the level of unobserved confounding at which the
bias in OPE estimates can raise concerns about the validity of selecting the
best policy among a set of candidates. As we illustrate, developing tools for
a meaningful sensitivity analysis is nontrivial: a naive bound yields
prohibitively conservative estimates that almost lose robustness certificates
for even neglible amounts of confounding, whereas our loss-minimization-based
bounds on policy values is informative.






\vspace{-8pt}
\subsection{Motivating example: managing sepsis patients}
\label{section:example}
\vspace{-2pt} Sepsis in ICU patients accounts for 1/3 of deaths in
hospitals~\cite{HowellDa2017}. Sepsis treatment decisions are made by a
clinical care team, including nurses, residents, and ICU attending physicians
and specialists~\citep{RhodesEvAl2017}. Difficulties of care often lead to
making decisions based off of imperfect information, leading to substantial
room for improvement. AI-based approaches provide an opportunity for optimal
automated management of medications, freeing the care team to allocate more
resources to critical cases. Automated approaches can manage important
medications for sepsis, including antibiotics and vasopressors, and decide to
notify the care team about when a patient should be placed on a mechanical
ventilator. Motivated by these opportunities, and the availability of ICU data
from MIMIC-III~\cite{JohnsonPoShLiFeGhMoSzCeMa2016}, several AI-based approaches
for sepsis management system have been proposed
\citep{FutomaLiSeBeClObHe2018,KomorowskiCeBaGoFa2018,RadhuKoCeSzGh2017}.

Due to safety concerns, new treatment policies need to be evaluated offline
before thorough online clinical validation.  Confounding, however, is a
serious issue in data generated from an ICU. Patients in emergency departments
often do not have an existing record in the hospital's electronic health
system, leaving a substantial amount of patient-specific information
unobserved in subsequent offline analysis. As a prominent example,
comorbidities that significantly complicate the cases of
sepsis~\cite{Brent2017} are often unrecorded. Private communication with an
emergency department physician revealed that \emph{initial} treatment of
antibiotics at admission to the hospital are often confounded by unrecorded
factors that affect the eventual outcome (death or discharge from the ICU).
For example, comorbidities such as undiagnosed heart failure can delay
diagnosis of sepsis, leading to slower implementation of antibiotic
treatments. More generally, there is considerable discussion in the medical
literature on the importance of quickly beginning antibiotic treatment, with
frequently noted concerns about confounding, as these discussions are largely
based on off-policy observational data collected from
ICUs~\cite{SeymourGePrFriwPhLeOsTeLe2017,SterlingMiPrPuJo2015}. Given the
recent interest in balancing early treatment with risks of over-prescription,
treatment regimes for antibiotics are of particular interest.

We consider a scenario where one wishes to evaluate between two automated
policies that differ only in initially \emph{avoiding}, or \emph{prescribing}
antibiotics, and otherwise acts optimally. The latter is often considered a
better treatment for sepsis, as it is caused by an infection. In this example,
unobserved factors most critically effect the first decision on prescribing
antibiotics upon arrival; since the care team is highly trained for treating
sepsis, we assume they follow standard protocols based on observed vitals
signs and lab measurements in subsequent time steps. In what follows, we
assess the impact of confounding factors discussed above on OPE of automated
policies, and provide certificates of robustness that guarantee gains over
existing policies.

\section{Related Work}
\label{section:related-work}

The majority of OPE methods for batch reinforcement learning rely on
sequential ignorability (though often unstated).
There is an extensive body of work for off-policy policy evaluation and
optimization under this assumption, including doubly robust
methods~\cite{jiang2015doubly,thomas2016data} and recent work that provides
semiparametric efficiency bounds~\cite{kallus2019double}; often the behavior
policy (conditional distribution of decisions given states) is assumed to be
known. Notably, \citet{liu2018representation} highlights how estimation error
in the behavioral policy can bias value estimates,
and~\citet{nie2019learning,hanna2019importance} provides OPE estimators based
on an estimator of the behavior policy. When sequential ignorability doesn't
hold, the expected cumulative rewards under an evaluation policy cannot be
identified from observable data. All of the above estimators are biased in the
presence of unobserved confounding, since neither the outcome model nor the
importance sampling weights can correct for the effect of the unobserved
confounder.

The do-calculus and its sequential backdoor criterion on the associated
directed acyclic graph \citep{Pearl09} also gives identification results for
OPE. Like sequential ignorability, this preclude the existence of unobserved
confounding variables. Therefore, methods that assume the sequential backdoor
criterion will be biased in their presence.


We study the effects of unobserved confounding on OPE in sequential decision
making problems, deriving bounds on the performance of the evaluation policy
when sequential ignorability is relaxed. For problems where only one decision
is made, a variety of methods developed in the econometrics, statistics, and
epidemiology literature estimate bounds on treatment effects and expected
rewards. \citet{Manski90} developed bounds that only assume bounded rewards,
though they are too conservative to identify whether one action is superior to
another. Then, \citet{Manski90} and other works posit models that bound the
effect of unobserved confounding on the outcome
\citep{Robins00,BrumbackHeHaRo04}, or---like ours---on the actions taken by
the behavior policy \citep{CornfieldHaHaLiShWy59, RosenbaumRu83,
  Imbens03}. Recent work studied approaches that can apply to heterogeneous
treatment effects~\cite{YadlowskyNaBaDuTi18,KallusMaZh2018}, policy
evaluation~\cite{JungShFeGo18}, and policy optimization~\cite{KallusZh2018}.

In sequential decision making settings, \citet{ZhangBa2019} derived partial
identification bounds on policy performance with limited restrictions on the
influence of the unobserved confounder on observed decisions, much like the single decision work of \citet{Manski90}, which they use to guide online RL algorithms. Unfortunately, these bounds are quite conservative for use only in OPE.
\citet{Robins00,Robins04,BrumbackHeHaRo04} instead posit a
model for how the confounding bias in each time step affects the outcome of
interest and derive bounds under this model: their work is motivated by
potential confounding in the effects of dynamic treatment regimes for HIV
therapy on CD4 counts in HIV-positive men. Our work is complementary to these
in that we instead assume a model for how the unobserved confounder affects
the actions taken by the behavior policy.

\section{Formulation}
\label{section:formulation}

Notation conventions vary substantially in the diverse set of communities
interested in learning from (sequential) observational data. In this paper, we
use the potential outcomes notation to make explicit which sequence of actions
we wish to evaluate versus which sequence of actions were actually observed. In
this approach, we posit all potential states and
rewards exist for each possible sequence of actions, but we only observe the one corresponding to the actions taken
(also known as partial, or bandit feedback), making the other potential states and rewards counterfactual. Literature in batch off policy reinforcement
learning almost always assumes sequential ignorability, in which case the
distribution of \emph{potential} states and rewards are independent of the
action taken by the behavior policy, conditional on the observed
history. This allows us to consistently estimate counterfactuals simply based on
observed outcomes. However, since our aim is to consider the impact of
hypothetized confounding, clarifying the difference between the potential and observed states and rewards is cumbersome, but important.

We focus on domains modeled by episodic stochastic decision processes with a
discrete set of actions. Let $\mc{A}_t$ be a finite action set of actions
available at time $t = 1, .., T$. Denote a sequence of actions
$a_1 \in \mc{A}_1, .., a_T \in \mc{A}_T$ by $a_{1:T}$ (and similarly
$a_{t:t'}$ for arbitrary indices $1\le t \le t' \le T$, with the convention
$a_{1:0} = \varnothing$).  For any sequence of actions $a_{1:T}$, let
$S_t(a_{1:t-1})$ and $R_t(a_{1:t})$ be the state and reward at time $t$. A
state can be a scalar or a vector of discrete- or continuous-valued features; in our experimental settings we consider continuous-valued states.
$Y(a_{1:T}) \defeq \sum_{t=1}^T \disc^{t-1} R_t(a_{1:t})$ is the corresponding
discounted sum of rewards. We denote by
$W(a_{1:T}) = (S_1, .., S_T(a_{1:T-1}), R_1(a_1), .., R_T(a_{1:T}))$ all
potential outcomes (over rewards and states) associated with the action
sequence $a_{1:T}$. Any sum $\sum_{a_{1:t}}$ over action sequences is taken
over all $a_{1:T} \in \mc{A}_1 \times \cdots \times \mc{A}_T$.

In the off-policy setting, we observe sequences of actions $A_1, .., A_T$
generated by an unknown behavior policy $\pi_1, .., \pi_T$. Let $H_t$ denote
the observed history until time $t$, so that $H_1 \defeq S_1$, and for
$t = 2, .., T$,
  $H_t \defeq (S_1, A_1, S_2(A_1), A_2, .., S_t(A_{1:t-1}))$.
As a notational shorthand, for any fixed sequence of actions $a_{1:T}$,
denote an instantiation of the observed history following the action
sequence by $H_t(a_{1:t-1})$, so that $H_1(a_{1:0}) \defeq H_1 = S_1$, and
  for $t = 2, .., T$,
$H_t(a_{1:t-1}) = (S_1, A_1 = a_1, S_2(a_1), .., A_{t-1} = a_{t-1},
S_t(a_{1:t-1}))$.
Let $\mc{H}_t$ be the set over which this history takes values.

When there is no unobserved confounding, $A_t \sim \pi_t(\cdot \mid H_t)$
since actions are generated conditional on the observed history $H_t$. When
there is \textit{unobserved confounding $U_t$}, the behavioral policy draws
actions $A_t \sim \pi_t(\cdot \mid H_t, U_t)$, and we denote by
$\pi_t(\cdot \mid H_t)$ the conditional distribution of $A_t$ given only the
observed history $H_t$, meaning we marginalize out the unobserved confounder
$U_t$.
For simplicity, we assume that previously observed rewards are included in the
states, so for $s < t$, $R_s(A_{1:s})$ is known given $H_t$ the history.
We define
$Y_t(a_t) \defeq Y(A_{1:t-1}, a_t, A_{t+1:T})$ as a shorthand: semantically 
this means the sum of rewards which matches a trajectory of executed actions 
on all but one action, where on time step $t$ action $a_t$ is taken. Note that 
since $a_t$ may not be identical to the taken action  $A_{t}$,  
the resulting expression for $Y$ represents a potential outcome. 


Our goal is to bound the performance of an evaluation policy
$\bar{\pi}_1, .., \bar{\pi}_T$ in a confounded sequential off-policy
environment.
Let
$\bar{A}_t \sim \bar{\pi}_t(\cdot \mid \bar{H}_t)$ be the actions generated by
the evaluation policy at time $t$, where we use
$\bar{H}_t \defeq (S_1, \bar{A}_1, S_2(\bar{A}_1), \bar{A}_2, ..,
S_t(\bar{A}_{1:t-1}))$ and
$\bar{H}_t(a_{1:t-1}) \defeq (S_1, \bar{A}_1 = a_1, S_2(a_1), \bar{A}_2 = a_2,
.., S_t(a_{1:t-1}))$ to denote the history under the evaluation policy,
analogously to the shorthands $H_t, H_t(a_{1:t-1})$; $\bar{H}_t$ are
mathematical constructs, as they are never observed in the behavioral data. We
are interested in statistical estimation of the expected cumulative reward
$\E[Y(\bar{A}_{1:T})]$ under the evaluation policy, which we call the
\emph{performance} of the evaluation policy (aka $V^{\bar{\pi}}$ in batch RL).
Throughout, we assume $\pi_t(a_t \mid H_t) > 0$ whenever
$\bar{\pi}_t(a_t \mid \bar{H}_t) > 0$, for all $t$ and $a_t$, and almost every
$H_t$,: in other words, overlap holds with respect to the conditional
distributions over actions given only the histories between the behavior
policy and the evaluation policy.

We now state the sequential ignorability assumption in terms of the
relationship between actions and potential outcomes (see
e.g~\cite{Robins86,Robins04,Murphy03}).
\begin{definition}[Sequential Ignorability]
  \label{def:seq}
  We say that a policy satisfies sequential ignorability if for all
  $t = 1, .., T$, conditional on the history generated by the policy, the
  action generated by the policy is independent of the potential outcomes
  $R_t(a_{1:t}), S_{t+1}(a_{1:t}), R_{t+1}(a_{1:t+1}), S_{t+2}(a_{1:t+1}),
  ..,$ $S_{T}(a_{1:T-1}), R_T(a_{1:T})$ for all
  $a_{1:T} \in \mc{A}_1 \times \cdots \mc{A}_T$.
\end{definition}

Sequential ignorability is a natural condition required for the evaluation
policy to be well-defined: any additional randomization used by the evaluation
policy $\bar{\pi}_t(\cdot \mid \bar{H}_t)$ cannot depend on unobserved
confounders. We assume that the evaluation policy always satisfies this
assumption.

\begin{assumption}
  \label{assumption:cand-seq-ig} The evaluation policy satisfies sequential ignorability (Definition~\ref{def:seq}). 
\end{assumption}

Off-policy policy evaluation fundamentally requires counterfactual reasoning
since we only observe the state evolution $S_t(A_{1:t-1})$ and rewards
$R_t(A_{1:t})$ corresponding to the actions made by the behavioral policy.
The canonical assumption in batch off-policy reinforcement learning is 
that sequential
ignorability \emph{holds for the behavior policy}. We now briefly review how this 
allows identification (and thus, accurate estimation) of $\E[Y(\bar{A}_{1:T})]$, 
the value of the evaluation policy.

Because we only observe potential outcomes $W(A_{1:t})$ evaluated at the
actions $A_{1:t}$ taken by the behavior policy $\pi_t$, we need to express
$\E[Y(\bar{A}_{1:T})]$ in terms of observable data generated by the behavioral
policy $\pi_t$. Sequential ignorability of both the behavior policy and
evaluation policy allows such counterfactual reasoning. The following identity
is standard; we give its proof in Section~\ref{section:proof-seq-ig} for
completeness. To ease notation, we write
\begin{equation}
  \label{eqn:is}
  \rho_t \defeq \frac{\bar{\pi}_t(A_t \mid \bar{H}_t(A_{1:t-1}))}{\pi_t(A_t \mid H_t)}.
\end{equation}
\begin{lemma}
  \label{lemma:seq-ig}
  Assume sequential ignorability (Definition~\ref{def:seq}) holds for both the 
  behavior and evaluation policy. Then, $\E[Y(\bar{A}_{1:T})]
    = \E[ Y(A_{1:T})
    \prod_{t=1}^T \rho_t
    ].$
\end{lemma}
\vspace{-4pt}
The RHS is called the importance sampling formula.


\section{Bounds under unobserved confounding}
\label{section:bounds}

Despite the advantageous implications, it is often unrealistic to assume that
the behavior policy $\pi_t$ satisfies sequential ignorability
(Definition~\ref{def:seq}). We now relax the sequential ignorability of the behavior
policy, and instead posit a model of bounded confounding for the behavior policy, then develop worst-case bounds on the evaluation policy performance
$\E[Y(\bar{A}_{1:T})]$ under this model. In addition to the observed state $S_t(A_{1:t-1})$
available in the data, we assume that there is an unobserved confounder
$U_t$ available only to the behavior policy at each time $t$. The behavior
policy observes the history $H_t$ and the unobserved confounder $U_t$, and
generates an action $A_t \sim \pi_t(\cdot \mid H_t, U_t)$. If $U_t$ contains
information about unseen potential outcomes, then sequential ignorability
(Definition~\ref{def:seq}) will fail to hold for the behavior policy.

Without loss of generality, let $U_t$ be such that the potential outcomes are
independent of $A_t$ when controlling for $U_t$ alongside the observed states.
Such an unobserved confounder always exists since we can define $U_t$ to be
the tuple of all unseen potential outcomes.
\begin{assumption}
  \label{assumption:confounder}
  For all $t = 1, .., T$, there exists a random vector $U_t$ such that
  conditional on the history $H_t$ generated by the behavior policy \textbf{and}
  $U_t$, $A_t \sim \pi_t(\cdot \mid H_t, U_t)$ is independent of the potential
  outcomes
  $R_t(a_{1:t}), S_{t+1}(a_{1:t}), R_{t+1}(a_{1:t+1}), S_{t+2}(a_{1:t+1}),
  ..,$ $S_{T}(a_{1:T-1}), R_T(a_{1:T})$ for all
  $a_{1:T} \in \mc{A}_1 \times \cdots \mc{A}_T$.
\end{assumption}
Identification of $\E[Y(\bar{A}_{1:T})]$ is impossible under arbitrary
unobserved confounding.  However, it is often plausible to posit that the
unobserved confounder $U_t$ has a limited influence on the decisions of the
behavior policy. When the influence of unobserved confounding on each action
is limited, we may expect OPE estimates that (incorrectly) assume sequential
ignorability may not be too biased.

Consider the following model of unobserved confounding for sequential decision
making problems, which bounds confounder's influence on the
behavior policy's decisions.
\begin{assumption}
  \label{assumption:bdd-conf}
  For $t=1, .., T$, there is a $\Gamma_t \ge 1$ satisfying
  \begin{align}
    \label{eqn:bdd-conf}
    \frac{\pi_t(a_t \mid H_t, U_t = u_t)}{\pi_t(a_t' \mid H_t, U_t = u_t)}
    \frac{\pi_t(a_t' \mid H_t, U_t = u_t')}{\pi_t(a_t \mid H_t, U_t = u_t')}
    \le \Gamma_t
  \end{align}
  for any $a_t, a_t' \in \mc{A}_t$, almost surely over $H_t$, and $u_t, u_t'$, and sequential ignorability holds conditional on $H_t$ \textbf{and} $U_t$.
\end{assumption}
\noindent Our bounded unobserved confounding assumption~\eqref{eqn:bdd-conf} is a
natural extension of a classical model of confounding proposed
by~\citet{Rosenbaum02} for a single decision ($T=1$) to sequential problems.
When the action space is binary $\mc{A}_t = \{0, 1\}$, the above bounded
unobserved confounding assumption is equivalent~\citep{Rosenbaum02} to the
following logistic model
$\log \frac{\P(A_t = 1 \mid H_t, U_t)}{\P(A_t = 0 \mid H_t, U_t)} =
\kappa(H_t) + (\log \Gamma_t) \cdot b(U_t)$
for some measurable function $\kappa(\cdot)$ and a bounded measurable function
$b(\cdot)$ taking values in $[0, 1]$. 

In the sequential setting where $T > 1$, OPE is almost always unreliable even
under the aformentioned model. Effects of confounding can create
exponentially large (in the horizon $T$) over-sampling of large (or small)
rewards, introducing an extremely large, un-correctable bias. As an
illustration, consider applying OPE in the following simplified setting, where
there are no states.  Let $U \sim \mbox{Unif}(\{0, 1\})$ be a single
unobserved confounder, and consider the sequence of behavioral actions
$A_1,\dots,A_T \in \{0, 1\}$ each drawn conditionally on $U$, but independent
of one another, with the conditional distribution
$P(A_t = 1 \mid U=1) = \sqrt{\Gamma }/ (1 + \sqrt{\Gamma})$ and
$P(A_t = 1 \mid U=0) = 1 / (1 + \sqrt{\Gamma})$.  Let the outcome be
$Y(a_{1:T}) = U$ for all possible action sequences $a_{1:T}$. Although the
actions do not affect the outcome, in the observed data the likelihood of
observing $((A_t = 1)_{t=1}^T, Y = 1)$ is
$\Gamma^{T/2}/(2(1+\sqrt{\Gamma})^{T})$, whereas the likelihood of observing
$((A_t = 1)_{t=1}^T, Y = 0)$ is $1/(2(1+\sqrt{\Gamma})^{T})$. Therefore, even
in the limit of infinite observations, OPE will mistakenly estimate that
always taking $\bar{A}_t=1$ leads to better rewards than always taking
$\bar{A}_t = 0$.

Even in this toy example example where states don't exist and rewards don't
depend on actions, the effect of confounding is salient. The unobserved
confounder can make certain observed data samples exponentially more likely
than others, without the OPE algorithm being able to tell or correct for these
differences. This has important implications for off-policy policy selection
or optimization, where such systematic differences can lead to selection of a
poorly performing policy.
  


\section{Confounding in a single decision}
\label{section:one-decision}
\label{section:convex}

In many important applications, it is realistic to assume there is only a
single step of confounding at a known time step $t^*$. Under this assumption,
we outline in this section how we obtain a computationally and statistically
feasible procedure for computing a lower (or upper) bound on the value
$\E[Y(\bar{A}_{1:T})]$ of an evaluation policy $\bar{\pi}$. After introducing
precisely our model of confounding, we show in
Proposition~\ref{prop:single-like-ratio} how the evaluation policy value can
be expressed using likelihood ratios over potential outcomes that can be used
to relate the potential outcomes over observed (factual) actions with
counterfactual actions not taken. These likelihood ratios over potential
outcomes are unobserved, but a lower bound on the evaluation policy value can
be computed by minimizing over all feasible likelihood ratios that satisfy our
model of bounded confounding. Towards computational tractability, we derive a
dual relaxation that can be represented as a loss minimization procedure. 

We define the confounding model for when there is an unobserved confounding
variable $U$ that only affects the behavior policy's action at a single time
period $t\opt \in [T]$. For example, in looking at the impact of confounders on
antibiotics in sepsis management (Section~\ref{section:example}), it is
plausible to assume that while confounders may influence the first decision when the patient arrives, later treatment decisions are not impacted by 
unobserved confounders.
\begin{assumption}
  \label{assumption:single-confounder}
  For all $t \neq t\opt$, conditional on the history $H_t$ generated by the
  behavior policy, $A_t$ is independent of the potential outcomes 
  $R_t(a_{1:t}), S_{t+1}(a_{1:t}), R_{t+1}(a_{1:t+1}), S_{t+2}(a_{1:t+1}),
  ..,$ $S_{T}(a_{1:T-1}), R_T(a_{1:T})$ for all
  $a_{1:T} \in \mc{A}_1 \times \cdots \mc{A}_T$. For $t = t\opt$, there exists
  a random variable $U$ such that the same conditional independence holds only
  when conditional on the history $H_t$ \textbf{and $U$}.
\end{assumption}
Similar to Assumption~\ref{assumption:bdd-conf}, but now restricted to a single time step $t^*$, we assume the unobserved confounder has bounded influence on the behavior policy's action $A_{t\opt}$.
\begin{assumption}
  \label{assumption:single-bdd-conf}
  There is a $\Gamma \ge 1$ satisfying
  {\small
  \begin{align}
    \label{eqn:single-bdd-conf}
    \frac{\pi_{t\opt}(a_{t\opt} \mid H_{t\opt}, U = u)}{\pi_{t\opt}(a_{t\opt}' \mid H_{t\opt}, U = u)}
    \frac{\pi_{t\opt}(a_{t\opt}' \mid H_{t\opt}, U = u')}{\pi_{t\opt}(a_{t\opt} \mid H_{t\opt}, U = u')}
    \le \Gamma
  \end{align}
  } \hspace{-5pt} for any $a_{t\opt}, a_{t\opt}' \in \mc{A}_{t\opt}$, almost surely over
  $H_{t\opt}$, and $u, u'$.
\end{assumption}
Selecting the amount of unobserved confounding $\Gamma$ is a modeling task,
and the above confounding model's simplicity and interpretability makes it
advantageous for enabling modelers to choose a plausible value of $\Gamma$. As
in any applied modeling problem, the amount of unobserved confounding $\Gamma$
should be chosen with expert knowledge (e.g. by consulting doctors that make
behavioral decisions). In Section~\ref{section:experiments}, we give various
application contexts in which a realistic range of $\Gamma$ can be posited.
One of the most interpretable ways to assess the level of robustness to
confounding is via the \emph{design sensitivity} of the
analysis~\citep{Rosenbaum10}: the value of $\Gamma$ at which the bounds on the
evaluation policy's value crosses a landmark threshold (e.g. performance of
behavior policy or some known safety threshold).

We first show that a simple na\"ive lower bound on the evaluation policy
performance $\E[Y(\bar{A}_{1:T})]$ can be obtained by directly applying our
bounded confounding model~\eqref{eqn:single-bdd-conf} to adjust the weights of
an importance sampling
estimator. 
Details are provided in Section~\ref{section:proof-naive}.
\begin{lemma}
  \label{lemma:naive}
  Let Assumptions~\ref{assumption:cand-seq-ig},~\ref{assumption:single-confounder},~\ref{assumption:single-bdd-conf} hold.
  Then, we have
  {\small
  \begin{align}
    \label{eqn:naive}
    & \E[Y(\bar{A}_{1:T})]
    \ge 
    \E\Bigg[ Y(A_{1:T}) \times \prod_{t=1}^T \rho_t \\
    & \qquad \times \left(\Gamma \indic{Y(A_{1:T}) < 0}
    + \Gamma^{-1} \indic{Y(A_{1:T}) > 0}\right)
    \Bigg].\nonumber
  \end{align}
  }
\end{lemma}
\vspace{-9pt} However, the naive bound~\eqref{eqn:naive} is often
prohibitively conservative, as we concretely illustrate in
Section~\ref{section:experiments}.

Instead we derive a tighter bound on the evaluation policy performance
$\E[Y(\bar{A}_{1:T})]$ based on a constrained convex optimization formulation
over counterfactual distributions. Under
Assumption~\ref{assumption:single-bdd-conf}, 
the likelihood ratio between observed and unobserved distribution at $t\opt$
can at most vary by a factor of $\Gamma$. Recall that $W(a_{1:T})$ is the
tuple of all potential outcomes associated with the actions $a_{1:T}$. The
following observation is due to~\citet[Lemma 2.1]{YadlowskyNaBaDuTi18}.
\begin{lemma}
  \label{lemma:single-like-ratio}
  Under
  Assumptions~\ref{assumption:single-confounder},~\ref{assumption:single-bdd-conf},
  for all $a_{t\opt} \neq a_{t\opt}'$, the likelihood ratio over the tuple of
  potential outcomes $W \defeq \{W(a_{1:T})\}_{a_{1:T}}$ exists
  \begin{equation*}
    \mc{L}(\cdot; H_{t\opt}, a_{t\opt}, a_{t\opt}')
    \defeq \frac{dP_{W}(\cdot \mid H_{t\opt}, A_{t\opt} = a_{t\opt}')}{dP_{W}(\cdot \mid H_{t\opt}, A_{t\opt} = a_{t\opt})},
  \end{equation*}
  and for $\P_{W}(\cdot \mid H_{t\opt}, A_{t\opt} = a_{t\opt})$-a.s. all $ w, w'$
  \begin{equation}
    \label{eqn:single-ratio-bound}
    \mc{L}(w; H_{t\opt}, a_{t\opt}, a_{t\opt}') \le \Gamma \mc{L}(w'; H_{t\opt}, a_{t\opt}, a_{t\opt}').
  \end{equation}
\end{lemma}
\noindent We let $\mc{L}(\cdot; H_{t\opt}, a_{t\opt}, a_{t\opt}) \equiv
1$. Using these (unknown) likelihood ratios, we can express the value of the
evaluation policy, $\E[Y(\bar{A}_{1:T})]$.
\begin{proposition}
  \label{prop:single-like-ratio}
  Under
  Assumptions~\ref{assumption:cand-seq-ig},~\ref{assumption:single-confounder},~\ref{assumption:single-bdd-conf},
  {\small
 \begin{align*}
   & \E[Y(\bar{A}_{1:T})] \\
  & = \E\Bigg[
    \prod_{t=1}^{t\opt-1} \rho_t  \sum_{a_{t\opt}, a_{t\opt}'}
   \bar{\pi}_{t\opt}(a_{t\opt} \mid \bar{H}_{t\opt}(A_{1:t\opt-1}))
    \pi_{t\opt}(a_{t\opt} \mid H_{t\opt}) 
       \\
   & \times \E\Big[
     \mc{L}(W; H_{t\opt}, a_{t\opt}, a_{t\opt}')
    Y(A_{1:T}) \prod_{t=t\opt+1}^T  \rho_{t} ~\Big|~ H_{t\opt}, A_{t\opt}=a_{t\opt}
    \Big]
     \Bigg].
 \end{align*}
 }
\end{proposition}
\vspace{-10pt} 
The proof is given in Section~\ref{section:proof-single-like-ratio}.

Proposition~\ref{prop:single-like-ratio} implies a natural
bound on the evaluation policy value $\E[Y(\bar{A}_{1:T})]$ under bounded unobserved
confounding. Since the likelihood ratios
$\mc{L}(\cdot; \cdot, a_{t\opt}, a_{t\opt}')$ are fundamentally unobservable
due to their counterfactual nature, we take a worst-case approach over all
likelihood ratios that satisfy condition~\eqref{eqn:single-ratio-bound}, and
derive a bound that only depend on observable distributions. Towards this
goal,  define  {\small
\begin{align}
  & \mathfrak{L}
    \defeq \big\{
    L: \mc{W} \times \mc{H}_{t\opt} \to \R_+ \mid
    L(w; H_{t\opt}) \le \Gamma L(w'; H_{t\opt})  \nonumber \\
  &  \mbox{a.s. all}~w, w',~\mbox{and}~
    \E[L(W; H_{t\opt}) \mid H_{t\opt}, A_{t\opt} = a_{t\opt}] = 1
    \big\}.
    \label{eqn:single-ratio-set}
\end{align}
} \hspace{-5pt} Taking the infimum over the inner expectation in the expression derived in
Proposition~\ref{prop:single-like-ratio}, and noting that it does not
depend on $a_{t\opt}'$, define
{\small
\begin{align*}
  & \eta\opt(H_{t\opt}; a_{t\opt}) \defeq \\
  & \inf_{L \in \mathfrak{L}}
  \E\left[
    L(W; H_{t\opt})
    Y(A_{1:T})
    \prod_{t=t\opt+1}^T \rho_{t} ~\Big|~ H_{t\opt}, A_{t\opt} = a_{t\opt}
    \right].
\end{align*}
} \hspace{-7pt} Since the above optimization is over infinite-dimensional
likelihoods, it is difficult to compute. We use functional convex duality to
derive a dual relaxation that can be computed by solving a \emph{loss
  minimization} problem over any well-specified model class. This allows us to
compute a meaningful lower bound to $\E[Y(\bar{A}_{1:T})]$ even when rewards
and states are continuous, by simply fitting a model using standard supervised
learning methods. For $(s)_+ = \max(s, 0)$ and $(s)_{-} = - \min(s,0)$, define
the weighted squared loss
  $\loss_{\Gamma}(z)
  \defeq \half (\Gamma (z)_-^2 + (z)_+^2)$.
  \begin{theorem}
  \label{theorem:loss-min}
  Let
  Assumptions~\ref{assumption:cand-seq-ig},~\ref{assumption:single-confounder},~\ref{assumption:single-bdd-conf}
  hold.  If
  $\E[ Y(A_{1:T})^2 \prod_{t=t\opt+1}^T \rho_{t}^2
  \mid A_{t\opt} = a_{t\opt}, H_{t\opt}] <\infty$ a.s., then
  $\eta\opt(H_{t\opt}; a_{t\opt})$ is lower bounded a.s. by the unique solution
  \begin{align*}
    & \kappa\opt(H_{t\opt}; a_{t\opt})
    = \argmin_{f(H_{t\opt})}
    ~~\E\Bigg[
    \frac{\indic{A_{t\opt} = a_{t\opt}}}{\pi_{t\opt}(a_{t\opt} \mid H_{t\opt})} \\
   & \qquad \qquad \times \loss_{\Gamma} \left(
    Y(A_{1:T})
    \prod_{t=t\opt+1}^T \rho_{t} - f(H_{t\opt})
    \right) \Bigg].
  \end{align*}
\end{theorem}
\noindent See Section~\ref{section:proof-loss-min} for the proof. From
Theorem~\ref{theorem:loss-min} and Proposition~\ref{prop:single-like-ratio},
our final lower bound on $\E[Y(\bar{A}_{1:T})]$ is given by
\begin{align}
  & \E\Bigg[
   \prod_{t=1}^{t\opt-1} \rho_t \sum_{a_{t\opt}}
  \bar{\pi}_{t\opt}(a_{t\opt} \mid \bar{H}_{t\opt}(A_{1:t\opt-1})) \nonumber \\
  &\qquad  \times (1-\pi_{t\opt}(a_{t\opt} \mid H_{t\opt})) \kappa\opt(H_{t\opt}; a_{t\opt})
    \Bigg] \nonumber \\
  & + \E\left[\pi_{t\opt}(A_{t\opt} \mid H_{t\opt}) Y(A_{1:T}) \prod_{t=1}^T \rho_t \right].
    \label{eqn:final-bound}
\end{align}

Our approach yields a loss minimization problem for each possible action,
where the dimension of this supervised learning problem is that of the
observed history $H_{t^*}$ generated from the behavior policy.
If confounding occurs early in the process ($t\opt \approx 1$), the space of
possible histories is small and this learning problem becomes easier. This is
the scenario for the domains we consider in our experiments.

In cases where there is low, yet sufficient, overlap, WIS can dramatically reduce variance, at the cost of increased bias, with respect to the usual IS estimator. While our approach uses the IS to adjust for the differences between the behavior and evaluation policy, adjusting the bound in \eqref{eqn:final-bound} to use WIS, instead, is straightforward. Altering the importance reweighting inside the loss function for $\kappa\opt$ to be normalized, like WIS, warrants further investigation.

\citet{YadlowskyNaBaDuTi18} takes a similar approach to bound the effect of
confounding on treatment effects when there is only one action taken. Our
approach allows for comparing sequences of actions derived according to an
evaluation policy, by adjusting for the way actions in all time steps depend
on the current states and history, and effect future states and rewards. One
notable challenge that only occurs in sequential problems is adjusting for
actions that occur after the confounded decision at time $t\opt$; these
actions depend on the confounded decision through the history generated.  A
natural approach is to individually bound the potential
outcomes $\E[Y(\bar{A}_{1:t\opt-1}, a_{t\opt:T})]$ for all $a_{t\opt:T}$,
where each bound is given by a loss minimization problem.  Under this
approach---which is analogous to that of~\citet{YadlowskyNaBaDuTi18} in the
single time step---computing a lower bound to $\E[Y(\bar{A}_{1:T})]$ requires
$\prod_{t=t\opt}^T |\mc{A}_t|$ loss minimization problems, making it
statistically and computationally intractable when $t\opt$ is small
(e.g. $t\opt = 1$ in our sepsis example). Instead, we consider averaged
outcomes $\E[Y(\bar{A}_{1:t\opt-1}, a_{t\opt}, \bar{A}_{t\opt+1:T})]$ in
Theorem~\ref{theorem:loss-min}, which allows us to obtain a lower bound on
$\E[Y(\bar{A}_{1:T})]$ by only solving $|\mc{A}_{t\opt}|$ loss minimization
problems.



\vspace{-10pt}
\paragraph{Consistency} We now show that an empirical approximation to our
loss minimization problem yields a consistent estimate of
$\kappa\opt(\cdot)$. We require the following standard overlap assumption,
which states that actions cannot be too rare under the behavior policy,
relative to the evaluation policy.
\begin{assumption}
  \label{assumption:overlap}
  There is $C \in (1, \infty)$ s.t. $\forall t$, $\rho_t \le C$, a.s..
\end{assumption}
Since it is not feasible to optimize over the class of all functions
$f(H_{t\opt})$, we consider a parameterization $f_{\theta}(H_{t\opt})$ where
$\theta \in \R^d$. We provide provable guarantees in the simplified setting
where $\theta \mapsto f_{\theta}$ is linear, so that the loss minimization
problem is convex. That is, we assume that $f_{\theta}$ is represented by a
finite linear combination of some arbitrary basis functions of $H_{t\opt}$. As
long as the parameterization is well-specified so that
$\kappa\opt(H_{t\opt}; a_{t\opt}) = f_{\theta\opt}(H_{t\opt})$ for some
$\theta\opt \in \Theta$, an empirical plug-in solution converges to
$\kappa\opt$ as the number of samples $n$ grows to infinity. We let
$\Theta \subseteq \R^d$ be our model space; our theorem allows
$\Theta = \R^d$.

In the below result, let $\what{\pi}_t(a_t \mid H_t)$ be a consistent
estimator of $\pi_t(a_t\mid H_t)$ trained on a separate dataset $\mc{D}_n$
with the same underlying distribution; such estimators can be trained using
sample splitting and standard supervised learning methods. Define the set
$S_{\epsilon}$ of $\epsilon$-approximate optimizers of the empirical plug-in
problem
{\small
\begin{align*}
  \min_{f(H_{t\opt})}
  \what{\E}_n\left[
  \frac{\indic{A_{t\opt} = a_{t\opt}}}{\what{\pi}_{t\opt}(a_{t\opt} \mid H_{t\opt})}
  \loss_{\Gamma} \left(
  Y(A_{1:T})
  \prod_{t=t\opt+1}^T \what{\rho}_{t} - f(H_{t\opt})
  \right) \right],
\end{align*}
}
\hspace{-5pt} where $\what{\E}_n$ is the empirical distribution on the data statistically independent from
$\mc{D}_n$, and
\begin{equation*}
  \what{\rho}_{t} \defeq \frac{\bar{\pi}(A_t \mid
    \bar{H}_t(A_{1:t-1}))}{\what{\pi}_t(A_t \mid H_t(A_{1:t-1}))}.
\end{equation*}
We assume we observe i.i.d. episodes, and that each episode (unit) does not
affect one another, so the observed cumulative reward is the evaluation of the
potential outcome at the observed action sequence, $Y(A_{1:T})$. 
We prove the below result in
Section~\ref{section:proof-single-consistency}.
\begin{theorem}
  \label{theorem:consistency}
  Let
  Assumptions~\ref{assumption:cand-seq-ig},~\ref{assumption:single-confounder},
  ~\ref{assumption:single-bdd-conf},~\ref{assumption:overlap} hold, and let
  $\theta \mapsto f_{\theta}$ be linear such that
  $f_{\theta\opt}(\cdot) = \kappa\opt(\cdot, a_{t\opt})$ for some unique
  $\theta\opt \in \R^d$. Let $\E |Y(A_{1:T})|^4 < \infty$, and
  $\E[|f_{\theta}(H_{t\opt})|^4] < \infty$ for all $\theta \in \Theta$. If for
  all $t$, $\what{\pi_t}(\cdot | \cdot) \to \pi_t(\cdot | \cdot)$ pointwise
  a.s., $\what{\rho}_{t} \le 2C$, and 
  $(2C)^{-1} \le \what{\pi}_{t\opt}(a_{t\opt} | H_{t\opt}) \le 1$ a.s., then
  $\liminf_{n \to \infty} \mbox{dist}(\theta\opt, S_{\varepsilon_n}) \cp
  0~\forall\varepsilon_n \downarrow 0$.
\end{theorem}
\vspace{-2ex} Hence, under the hypothesis of
Theorem~\ref{theorem:consistency}, a plug-in estimator of the lower
bound~\eqref{eqn:final-bound} is consistent as $n \to \infty$.



\section{Experiments}
\label{section:experiments}

\begin{figure}[!tb]\centering
\includegraphics[width=0.85\linewidth]{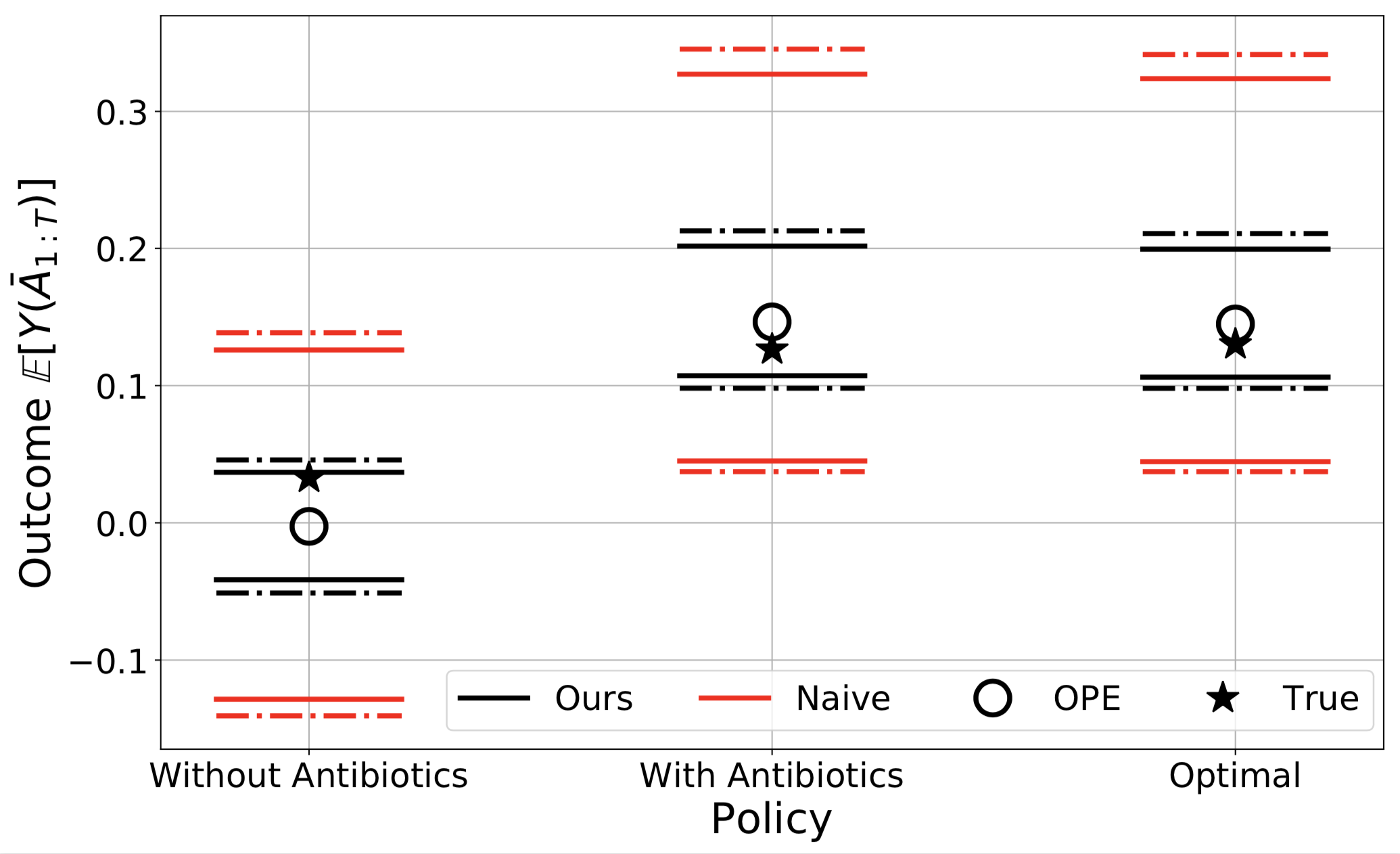}
\caption{Sepsis simulation. Data generation process with the level of confounding $\Gamma^\star =2.0$. Each policies' true value is shown with a start and a standard OPE estimate (ignoring confounding) is shown with an empty circle. 
Black lines show the estimated upper and lower bound on policy performance using our approach and red lines correspond to the na\"ive approach, both using $\Gamma=2.0$. Dashed lines represents $95\%$ quantile.}
\vspace{-2.0 mm}
\label{fig:sepsis}
\end{figure}

\begin{figure}[!tb]
  \centering  
  \begin{tabular}{cc}
    \includegraphics[width=.45\linewidth]{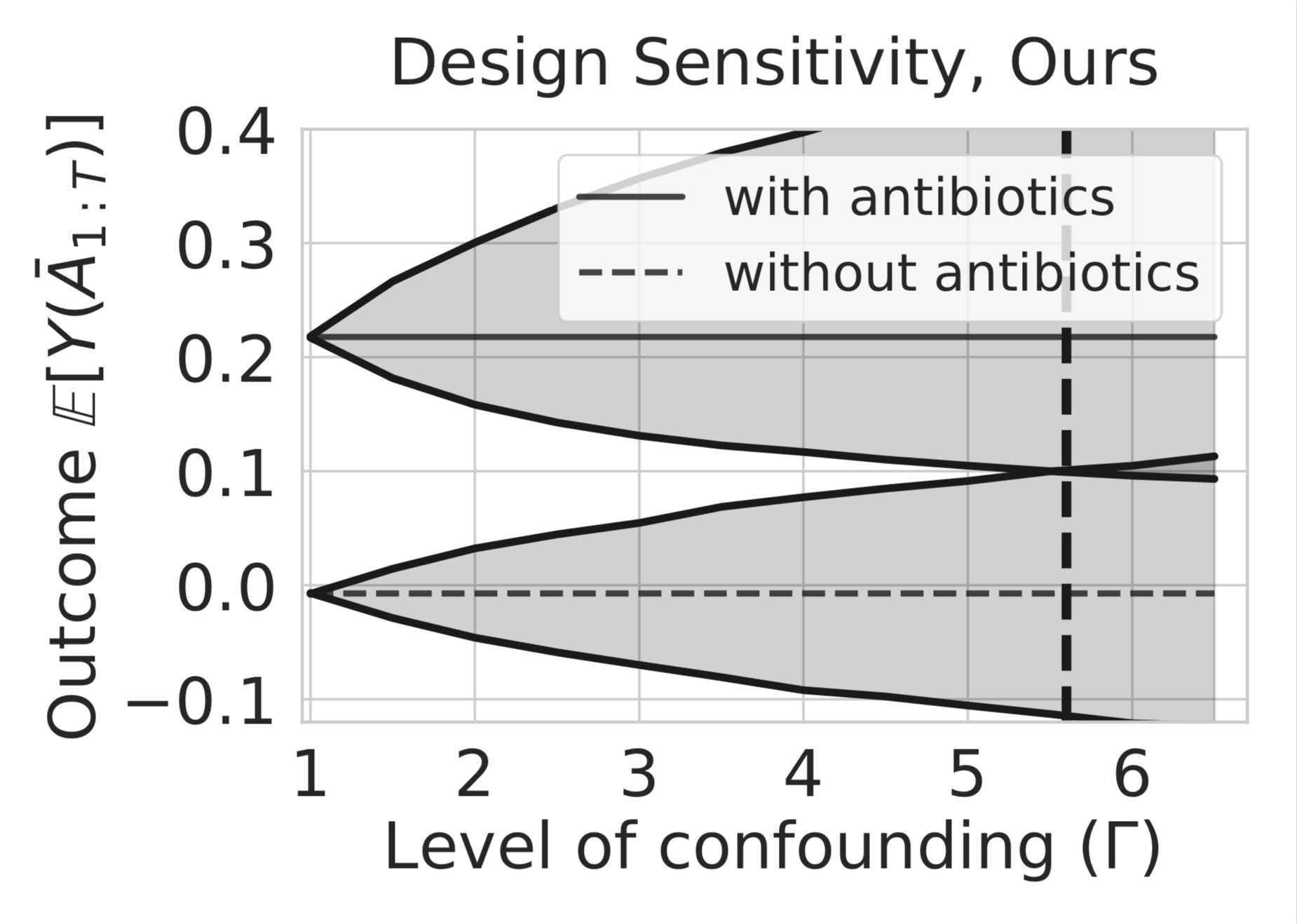}      
    &     
    \includegraphics[width=.45\linewidth]{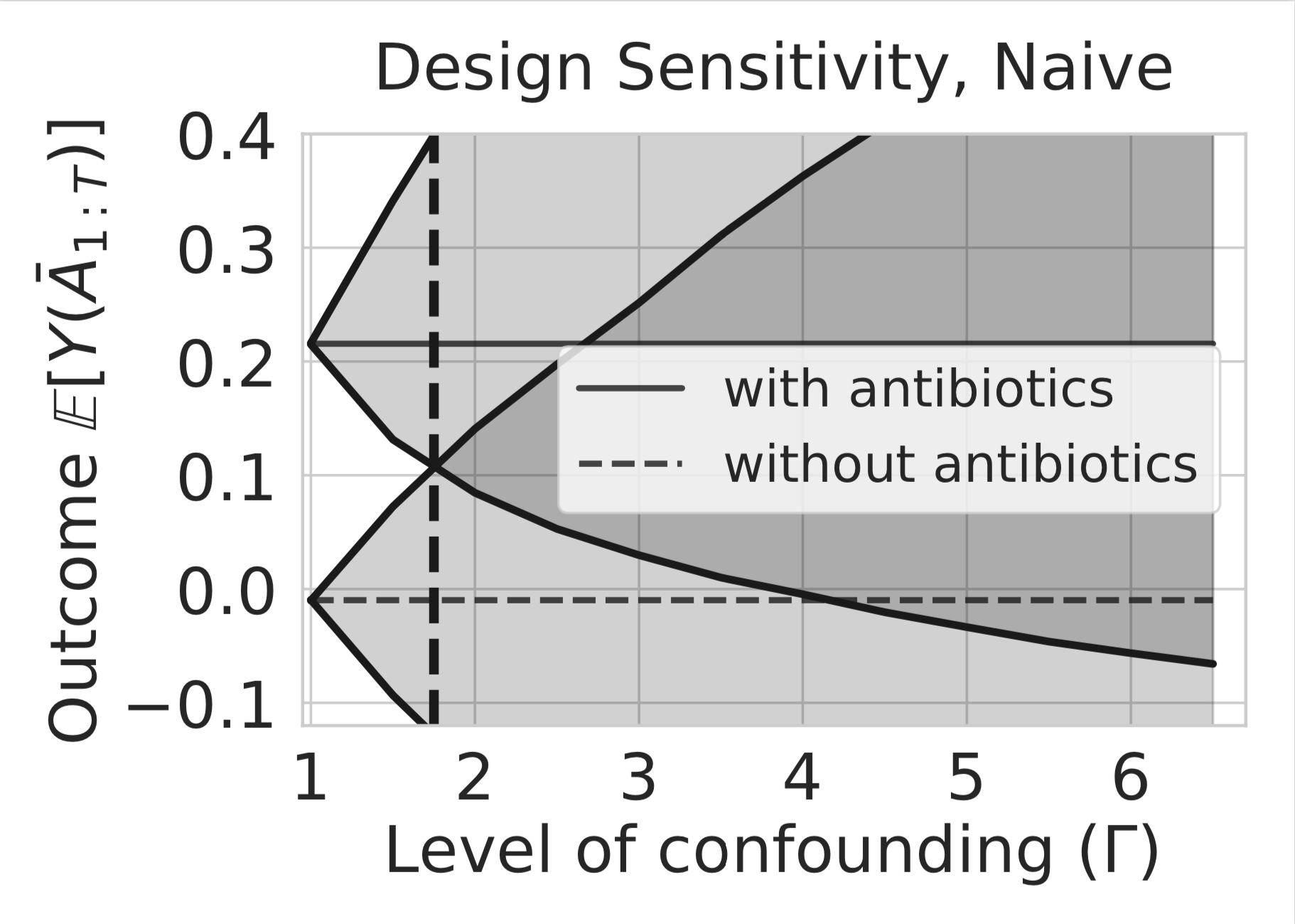} \\  
    {\small (a) Our approach} & {\small (b) Naive approach}
  \end{tabular}                                                  
  \caption{\label{fig:sepsis_design}Sepsis simulator design sensitivity. Data generation process with level of confounding $\Gamma^\star = 5$. Estimated lower and upper bound of two policies (with and without antibiotics) under (a) our approach with design sensitivity $5.6$ (b) naive approach with design sensitivity $1.75$.}         
  \vspace{-4 mm}
\end{figure}      

We illustrate how our approach can generate meaningful certificates of
robustness to unobserved confounding in realistic scenarios. We consider
selecting evaluation policies using off-policy evaluation methods, such as
comparing the expected performance of a new policy to an existing policy.
We empirically validate our method in sequential off-policy evaluation
problems where confounding is primarily an issue in only a single decision.
Since counterfactual outcomes are only known in simulations, we focus on
simulated healthcare examples motivated by two real OPE applications:
management of sepsis patients, and developmental interventions for autistic
children. We select these examples because they represent interesting cases
with existing simulators. As we argue shortly, it is plausible to assume that
unobserved confounding only affects a single decision in both settings.

The two scenarios characterize different problem regimes. The sepsis simulator
models discrete state space and the horizon of decision making ($T$) is
naturally multiple steps, and the autism management simulator models
continuous-valued states and horizon $T =2$. Our results demonstrate
scalability of our loss minimization approach in both discrete and continuous
settings, as well as short and medium horizons ($5\sim 10$). We observe that
beyond $10$ time steps, overlap becomes a problem, and statistical estimation
becomes challenging.


In both examples, we compare three
different approaches: standard OPE methods that (incorrectly) assume
sequential ignorability, the na\"ive bound~\eqref{eqn:naive}, and the bound
using our proposed loss minimization approach~\eqref{eqn:final-bound}. All the
code required to reproduce our experiments are available online at
\url{https://github.com/StanfordAI4HI/off_policy_confounding}. 
In both cases our approach provides informative bounds on the performance of
the evaluation policy, allowing reliable selection of policies even under
unobserved confounding. Compared to the na\"ive approach~\eqref{eqn:naive}
which is often prohibitively conservative, our methods allow certifying
robustness to much larger levels of confounding $\Gamma$.

\subsection{Managing sepsis for ICU patients}
As outlined in Section~\ref{section:example}, automated policies hold much
promise in management of sepsis in ICU patients. However, 
ICU observational data about sepsis patients 
may often 
lack information about important confounders, 
such as important unrecorded comorbidities that affect a clinician's initial
decision whether to administer antibiotics.
In subsequent time steps, we assume the (highly-trained) clinical care team
follows standard protocols based on vitals signs and lab measurements, and
hence their subsequent decisions are unconfounded. On the sepsis simulator
developed by \citet{OberstSo2019}, we illustrate how such confounders can bias OPE methods, and demonstrate that our worst-case approach can
allow reliable selection of candidate policies under confounding.

We consider a scenario where automated policies have been proposed using
existing medical knowledge, and we wish to evaluate their benefits relative to
the current standard of care.  We evaluate three different policies, all of
which only differ in their initial prescription of antibiotics, and otherwise
act optimally. The first policy, \emph{without antibiotics (WO)}, does not
administer antibiotics initially, whereas the second policy, \emph{with
  antibiotics (W)}, always administers antibiotics initially. For our last
policy, we follow \citet{OberstSo2019} and use the \emph{optimal} policy
learned by running policy iteration on this simulator---naturally this
procedure does not have confounding. We stress that our first two policies are
identical to the optimal policy after the initial time step. The true
performance of the with antibiotics (W) and optimal policy is quite similar,
and better than the without antibiotics (WO) policy (see Figure~\ref{fig:sepsis}).

To simulate unrecorded comorbidities that could introduce confounding, we
extract the randomness that governs state transitions into a confounding
variable so that the confounder is correlated with better state
transitions. In the first time step, we take the optimal action with respect
to all other options (vasopressors and mechanical ventilation), and administer
antibiotics with probability $\sqrt{\Gamma\opt}/(1+ \sqrt{\Gamma\opt})$ if the
confounding variable is large, and with probability
$1/(1 + \sqrt{\Gamma\opt})$ if the confounding variable is small. This
confounder satisfies Assumption~\ref{assumption:single-bdd-conf} with level
$\Gamma\opt$. Note that $\Gamma\opt$ is used in the data generation process,
but is unknown to the procedure used to estimate (bounds on) the evaluation
policy performance.  We run our method with varying levels of $\Gamma$, and
look at thresholds at which the bounds on the performance of evaluation
policies cross each other (which we refer to as the \emph{design
  sensitivity}).

To generate our observational data, we assume that the care team acts nearly
optimally, except for some randomness due to challenges in the ICU; this
guarantees overlap (Assumption~\ref{assumption:overlap}) with respect to the
optimal evaluation policy. In all but the first time step, we let the behavior
policy take the optimal next treatment action with probability $0.85$, and
otherwise switch the vasopressor status, independent of the confounders; this
guarantees that the assumption of single time step confounding (Assumption~\ref{assumption:single-confounder}) holds.

\citet{OberstSo2019}'s simulator state space consists of a binary indicator for diabetes, and four
vital signs \{heart rate, blood pressure, oxygen concentration and glucose
level\} that take values in a subset of \{very high, high, normal, low, very low\}; 
size of the state space is $|\mc{S}_t| = 1440$. There are three binary treatment
options for \{antibiotics, vasopressors, and mechanical ventilation\}, so that
the action space has cardinality $|\mc{A}_t|=2^3$.  In our experiments,
simulation continues either until at most $T = 5$ (horizon) time steps, death (reward
-1), or discharge (reward +1). Patients are discharged when all vital signs
are in the normal range without treatment. Patients die if at least three
vitals are out of the normal range. We refer the reader to
\url{https://github.com/clinicalml/gumbel-max-scm} for details regarding the
simulator.

\begin{figure}[!tb]
  \centering                                                     
  \begin{tabular}{cc}
    \includegraphics[width=.45\linewidth]{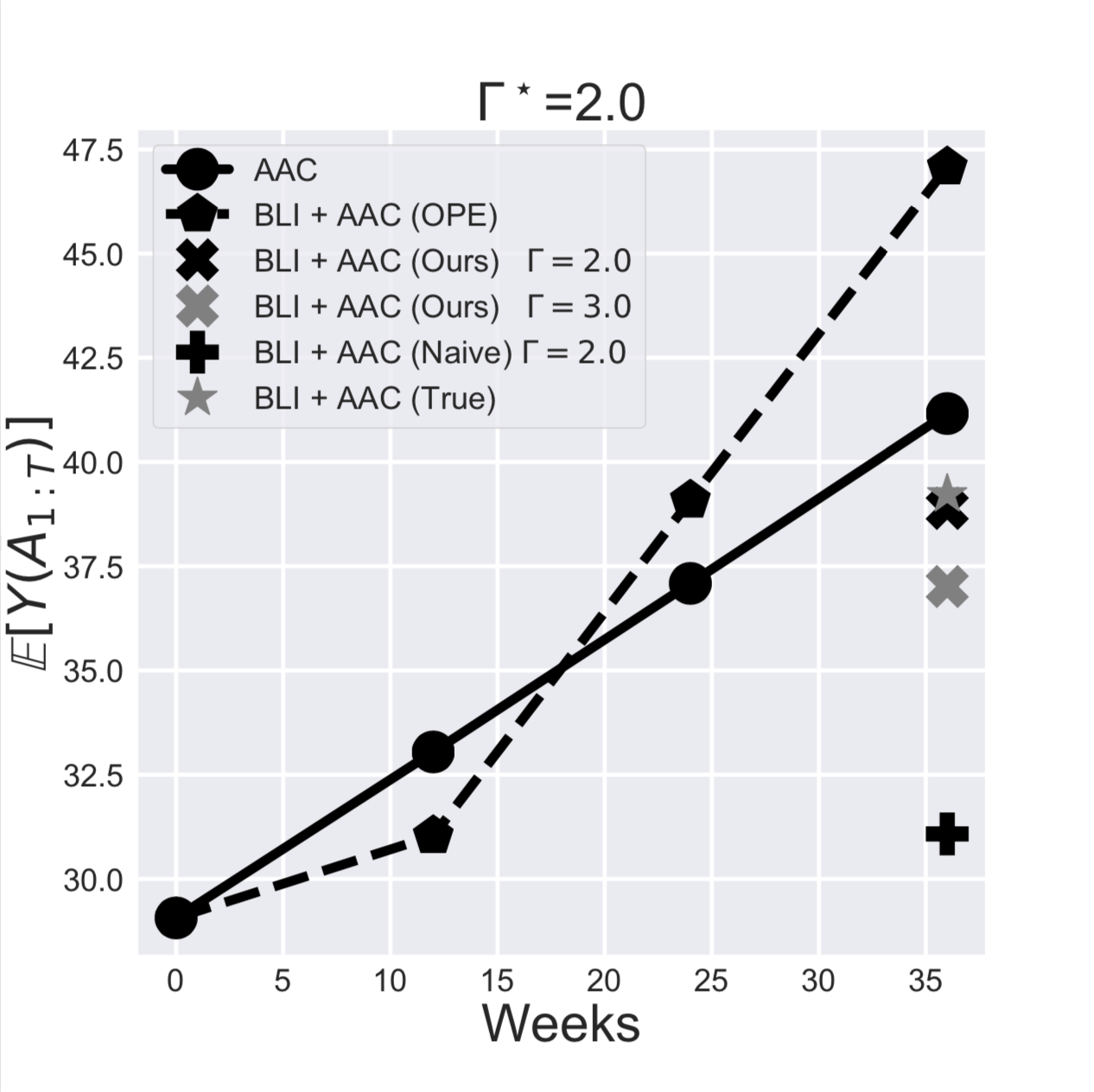}      
    &     
    \includegraphics[width=.45\linewidth]{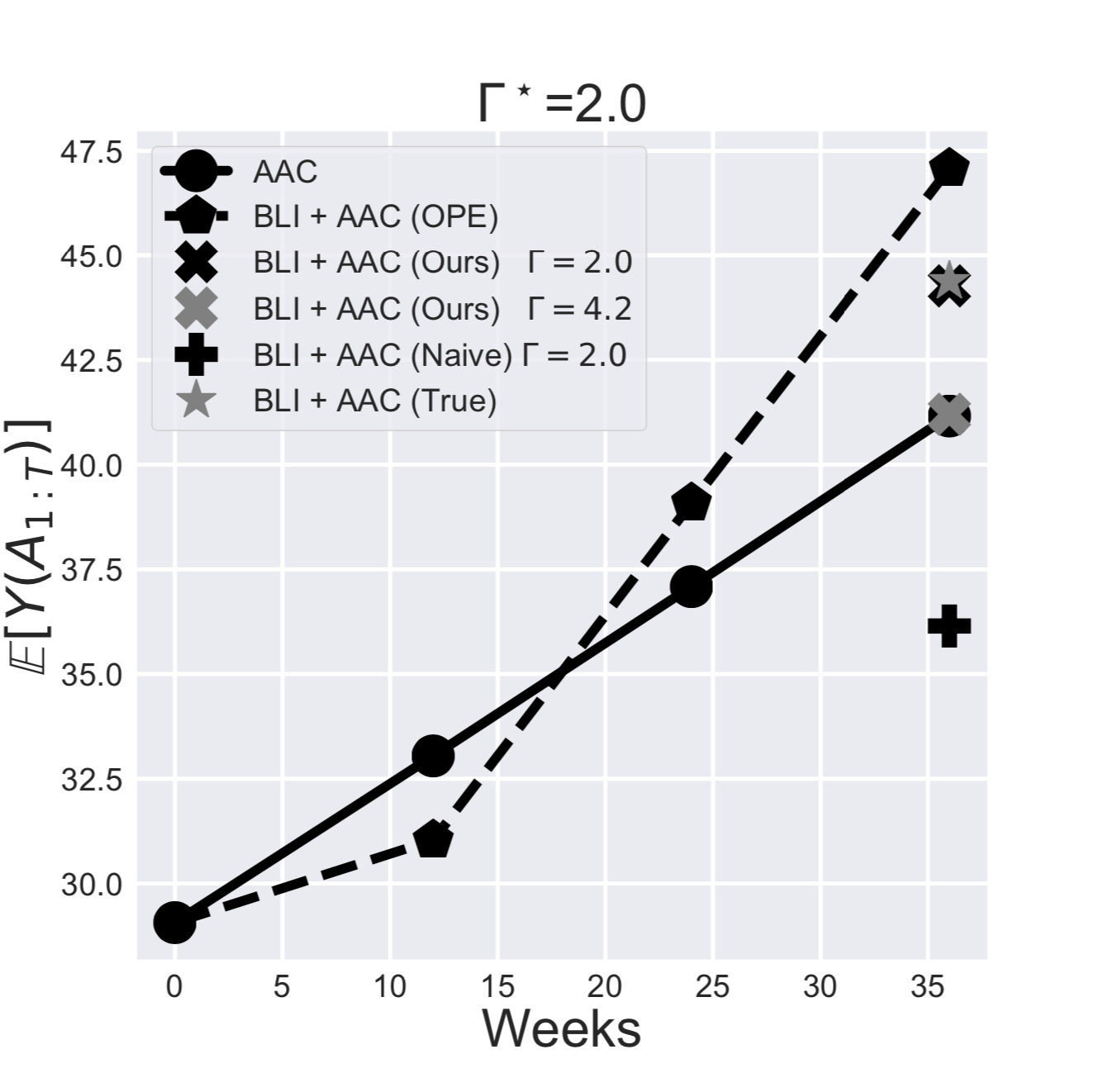} \\  
    {\small (a) Case I} & {\small (b) Case II}\\
  \end{tabular}                                                  
  \caption{\label{fig:autism}Autism simulation. Outcome of two different policies, confounded adaptive policy (BLI+AAC) and un-confounded non-adaptive policy (AAI). Data generation process with the level of confounding $\Gamma^\star=2.0$.}
\vspace{-3 mm}
\end{figure} 

\begin{figure}[!tb]\centering
\includegraphics[width=0.98\linewidth]{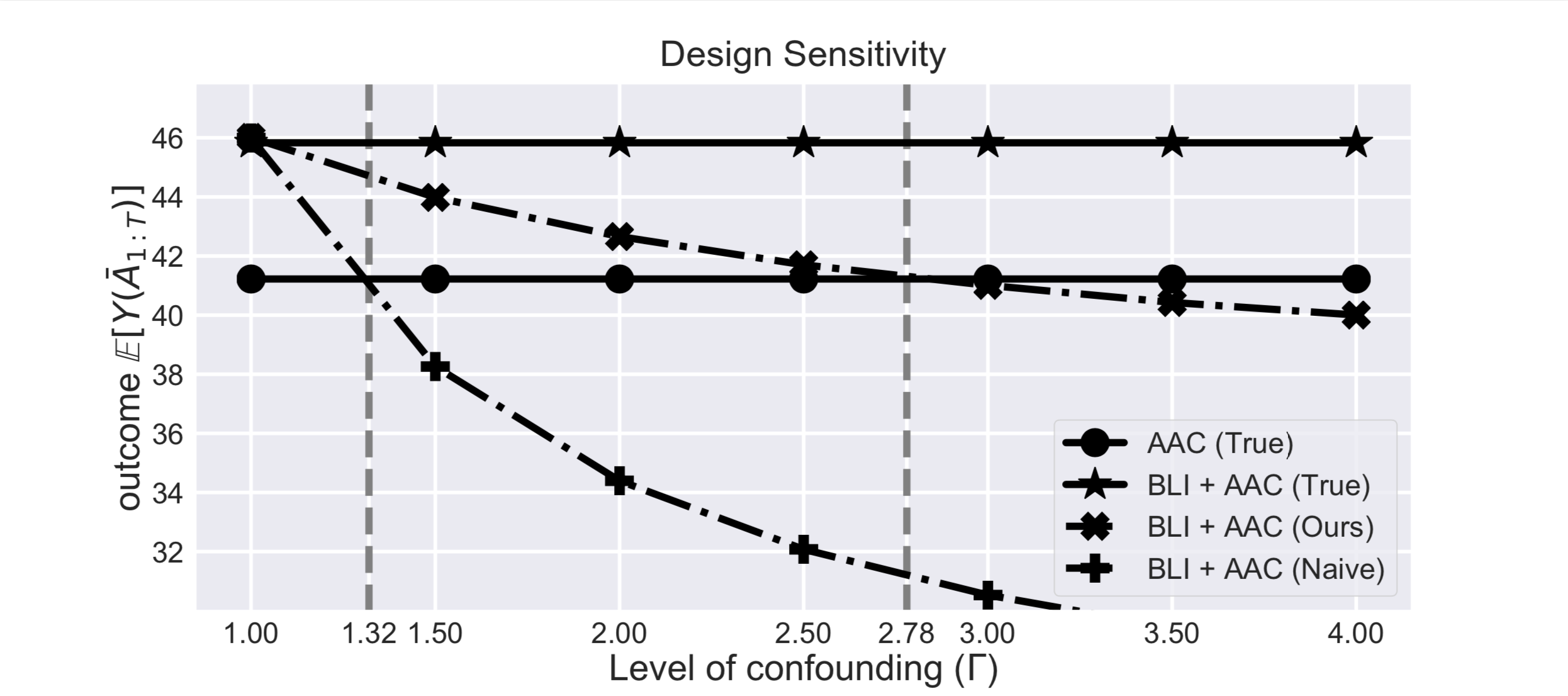}
\caption{Autism simulation design sensitivity. Data generation process with the level of confounding $\Gamma^\star =1.0$. True value of adaptive (BLI+AAC) and non-adaptive (AAC) policies along with estimated lower bound on outcome using our and naive approach}
\vspace{-6 mm}
\label{fig:autism_sen}
\end{figure}

We first consider when our approach happens to use the same confounding degree as what is present in the simulator, $\Gamma = \Gamma\opt$. 
Figure~\ref{fig:sepsis} plots the value of the three 
evaluation decision policies estimated using the
data generated with $\Gamma^\star=2.0$, which is a fairly small amount of 
confounding. %
Confounding leads standard OPE methods that assume sequential ignorability for the behavior policy 
to underestimate the peformance of 
the without antibiotics (WO) policy, and overestimate the performance 
of 
the with antibiotics (W) and optimal policies. This inflates the expected benefit of 
the W and optimal policies compared to the WO policy. 
The naive approach~\eqref{eqn:naive} results in very wide estimated intervals
over the potential policy performance, and therefore cannot be used to
reliably infer the superiority of W and optimal policy over WO even when
$\Gamma=2.0$. On the other hand, our proposed method certifies the robustness
of the benefit of immediately administering antibiotics; our lower bounds on
the performance of the W and optimal policies are better than the upper bound on the performance under the WO policy.

We next consider a much larger amount of confounding, generating the
observational data with $\Gamma^\star=5.0$. To explore the design sensitivity
of our method and our nai\"ive lower bound approach, we use a range of
$\Gamma$ values in our method. 
Figure~\ref{fig:sepsis_design} shows that for our method, the lower bound on
the performance of the W policy meets the upper bound on that of the WO policy
at $\Gamma=5.6$. In other words, our approach can reliably estimate that the W
policy is better than the WO policy up to assuming an amount of confounding
determined by $\Gamma=5.6$ when the true $\Gamma\opt=5.0$. In contrast, our
proposed na\"ive bound~\eqref{eqn:naive} has a a design sensitivity of
$\Gamma=1.75$, meaning the bounds quickly fails to be informative far below
the true amount of data confounding. Our method allow concluding that the W
policy is superior to the WO policy even when a substantial amount of
unobserved confounding exists in the initial decision.

\subsection{Communication interventions for minimally verbal children with autism}
We next consider another motivating scenario from healthcare, but one which
naturally involves continuous variables to demonstrate that our approach is
also able to compute reasonable lower bounds for such a case, while using function approximation. 

Minimally verbal children represent 25-30\% of children with autism, and often
have poor prognosis in terms of social functioning
\cite{RutterGrLo1967,AndersonOtLoWe2009}. We are interested in comparing
non-adaptive versus adaptive approaches that aim to improve spoken
communication, measured by the number of speech utterances. 
We introduce confounding using a simulator for autistic children developed
by~\citet{LuNaKaOsPeFaAl2016}, which models the data from a (real) sequential,
multiple assignment, randomized trial
(SMART)~\citep{KasariKaGoNiMaLaMuAl2014}. Despite their randomized trial,
\citet{KasariKaGoNiMaLaMuAl2014} note that very few randomized trials of these
interventions exist, and the number of individuals in these trials tends to be
small. It is therefore reasonable to think that in similar settings it would
be beneficial to use existing off-policy data to evaluate new intervention
protocols.

In the simulator there are two 
developmental interventions (actions): behavioral language interventions (BLI)
delivered by a therapist, and an augmented/alternative communication (AAC)
approach implemented with a speech generation device. There are two decision
points in the data generation process: week 0 and week 12. Number of speech 
utterances are measured at week 0, 12, 24 and 36: note that the action / intervention 
applied at week 12 persists from week 12 to the end of the process, which 
means this is a 2 time step decision problem. Here the outcome is modeled as a continuous variable representing the average number of speech utterances for a given patient. 

We consider a scenario where participants were recruited and randomly assigned
to the two treatment options initially (i.e.,
$A_1 \sim \mbox{Unif}(\left\{\mbox{BLI}, \mbox{AAC}\right\})$), and a recourse
action is taken after a follow-up visit after 12 weeks. Depending on the
progress of patients at Week 12, the clinician decides whether to switch to
AAC devices for children who started with BLI. Since this intervention
requires a specialized device---whose supply is limited---it is likely that
the clinicians assign AAC devices for whom it has a higher chance of being
effective. Such subjective assessments are likely based on the their
interaction with patients that contain partial, noisy information about the
final outcome, which are often not recorded properly. Therefore, while there
is confounding in the second decision ($t\opt=2$), its influence may be
appropriately bounded (i.e., Assumption~\ref{assumption:single-confounder} is
plausible). To simulate confounding, we expand the simulator to create
variables that partially influences the effectiveness of switching from BLI to
AAC, and use knowledge of this to alter the behavior policy decisions at Week
12. The resulting confounding satisfies our model of bounded confounding
(Assumption~\ref{assumption:single-confounder}) and is described in detail in
Appendix~\ref{app:autism}.

In our evaluations, we compare an adaptive policy (BLI + AAC) that starts with
BLI, and augments BLI with AAC at week 12 if the patient is a slow responder, against a non-adaptive policy that
uses AAC through the whole treatment.  We simulate two different settings where
the effect of switching to the AAC treatment varies; our simulation parameters
are within the suggested range of ~\citet{LuNaKaOsPeFaAl2016}'s
recommendations based on the SMART trial data. We note that OPE estimates for
the non-adaptive policy (AAC) is unbiased since observations for this outcome
are unconfounded. Our loss minimization for computing the lower bound using
$\kappa(a_{t\opt}, H_{t\opt})$ is done using a 4 layers neural network with
Relu activations, we use backpropogation with AdamOptimizer and weighted squared loss given
in Theorem~\ref{theorem:loss-min}. We use logistic regression to estimate the
behavior policy, note that this is the marginalized behavior policy since the
latent confounder is unobserved.

In Case I, we define the parameters such that the adaptive policy (BLI+AAC) is
worse the non-adaptive policy (AAC) (lower true outcome / performance). As
shown in Figure~\ref{fig:autism} (a), standard OPE approach overestimates the
outcome of the adaptive policy even given a mild level of confounding
$\Gamma^\star=2.0$, and would incorrectly suggest the BLI+AAC policy
outperforms the AAC policy. On the other hand, our lower bounds on
the adaptive policy computed using $\Gamma=2$ (recall the true confounding
amount is unknown to our approach) suggest the OPE estimates may be biased
enough to affect conclusions; the observed advantages of the adaptive policy
may be attributed solely to unobserved confounding, even under reasonable
values of confounding ($\Gamma = 2$). 


In Case II, we change the parameters so that the BLI+AAC policy is better than the
AAC policy, and again use a true amount of confounding of $\Gamma^\star=2.0$ 
in the data generation process. Standard OPE estimates 
again overestimate the outcome for the BLI+AAC policy (Figure~\ref{fig:autism}(b)). 
The na\"ive lower bound results in a conservative lower bound that would again 
indicate no conclusions can be drawn about the relative performance of 
BLI+AAC versus AAC. However, our method
can certify the superiority of the BLI+AAC policy when the level of confounding 
used in the computation of the lower bounds is up to $\Gamma=4.2$, 
thereby providing a case where our approach can provide useful certificates 
of benefit of a new decision policy under non-trivial levels of confounding.

Figure~\ref{fig:autism_sen} plots the design sensitivity of our method against
the na\"ive approach~\eqref{eqn:naive}, when there is in fact no confounding in
the data generation process ($\Gamma\opt = 1$). Compared to the naive approach
(design sensitivity is $\Gamma=1.32$), our method allows certifying robustness
of the finding---that the adaptive policy is advantageous---up to realistic
levels of confounding (design sensitivity is $\Gamma = 2.78$).



\section{Discussion}

In this work, we proposed methods for analyzing the sensitivity of OPE
methods to unobserved confounding in sequential decision making problems.  We
demonstrated how our approach can certify robustness of OPE in some settings,
or raise concerns about its validity based on sensitivity to unobserved
confounding. Our loss minimization method allows computing worst-case bounds
over our bounded unobserved confounding model, while adjusting for observed
features via importance sampling.

As a consequence, our estimators face the same challenges that standard
importance-sampling-based OPE methods face: high variance when there is little
overlap between the evaluation and behavior policy. In our experiments, importance sampling
was effective since we ensured that there was sufficient overlap and focused
on shorter horizons. In other settings, lack of overlap poses fundamental
difficulties in off policy evaluation, beyond issues with confounding, as
others have also noted~\cite{GottesmanJoKoFaSoDoCe2019}. Such challenges become
pronounced as the horizon $T$ or the importance sampling
weights~\eqref{eqn:is} become large. While stationary
importance sampling (SIS) can reduce variance, rewards under stationary
distributions (should they exist) are not appropriate for the problems
studied in this paper; SIS~\citep{hallak2017consistent,liu2018breaking,xie2019towards,liuoff} nevertheless still suffers high
variance when there is a lack of
overlap.
\citet{fujimoto2018off} and \citet{kumar2019stabilizing} suggest some
promising algorithmic approaches for only considering (and optimizing over) policies with sufficient overlap: 
while more work is needed, policies generated by these approaches would be more amenable to OPE, and should improve the statistical properties of our
method. 


It is natural to consider extending our single-decision confounding model to
settings where a handful of decisions (say 2-5) are affected by unobserved
confounding. Worst-case bounds on $\E[Y(\bar{A}_{1:T})]$ under such extensions
require solving optimization problems involving products of likelihood ratios
defined over different confounded time periods. Since these problems are
nonconvex, they require new approaches than the one we take here, which heavily depends on applying
convex duality.






\bibliography{bib}
\bibliographystyle{abbrvnat}



\newpage
\onecolumn
\appendix

\section{Proof of basic lemmas}
\label{section:helper-lemmas}

Before we give the proof of our main results, we give a set of essentially
standard lemmas that we build on in the rest of the paper. In the following,
we use a notational shorthand for (nested) expectations under observable
distributions: for all $1\le t\le T$ and $1\le t_1 \le t_2\le T$,
\begin{subequations}
  \label{eqn:shorthand}
\begin{align}
  & \E^t_{a_{t}}[X]
  \defeq \E[X \mid H_t, A_t = a_t] 
  ~~~\mbox{and}~\\
  & \E^{t_1:t_2}_{a_{t_1:t_2}}[X]
    \defeq \E^{t_1}_{a_{t_1}}[\E^{t_1 + 1}_{a_{t_1+1}}
    [\cdots \E^{t_2}_{a_{t_2}}[X] \cdots ]].
\end{align}
\end{subequations}
Similarly, we write for all $1\le t_0 \le t_1 \le t_2\le T$
\begin{subequations}
  \label{eqn:shorthand-two}
\begin{align}
  & \E^{t_2}_{a_{t_1:t_2}}[X]
  \defeq \E[X \mid H_t(A_{1:t_1-1}, a_{t_1:t_2-1}), A_{t_2} = a_{t_2}] 
  ~~~\mbox{and}~\\
  & \E^{t_1:t_2}_{a_{t_0:t_2}}[X]
    \defeq \E^{t_1}_{a_{t_0:t_1}}[\E^{t_1 + 1}_{a_{t_0:t_1+1}}
    [\cdots \E^{t_2}_{a_{t_0:t_2}}[X] \cdots ]].
\end{align}
\end{subequations}

The cumulative rewards $\E[Y(\bar{A}_{1:T})]$ under the candidate policy has
an alternate representation, which we draw on heavily in the rest of the
proofs. See Section~\ref{section:proof-cand-identity} for a derivation.
\begin{lemma}
  \label{lemma:cand-identity}
  If sequential ignorability (Assumption~\ref{assumption:cand-seq-ig}) holds
  for the evaluation policy $\bar{\pi}$, we have the identity
  \begin{equation*}              
    \E\left[Y(\bar{A}_{1:T})\right]
    = \sum_{a_{1:T}}  \E\left[
    Y(a_{1:T})
    \prod_{t=1}^T \bar{\pi}_t(a_t \mid \bar{H}_t(a_{1:t-1}))
    \right].
  \end{equation*}
\end{lemma}
\vspace{-10pt}
\noindent To ease notation, denote each integrand in the above sum by
\vspace{-10pt}
\begin{equation}
  \label{eqn:averaged-outcome}
  Y(a_{1:T}; \bar{\pi})
  \defeq Y(a_{1:T})
  \prod_{t=1}^T \bar{\pi}_t(a_t \mid \bar{H}_t(a_{1:t-1})).
\end{equation}
\vspace{-10pt}

We will also use the following two identities heavily. Recall that we denote
by $\outcomes \defeq \{W(a_{1:T})\}_{a_{1:T}}$, the tuple of all potential
outcomes, which takes values in $\mc{W}$. See
Section~\ref{section:proof-seq-ig-nested-exp} for a proof of the following
result.
\begin{lemma}
  \label{lemma:seq-ig-nested-exp}
  Let sequential ignorability (Assumption~\ref{assumption:cand-seq-ig}) hold
  for the behavioral policy $\pi$ in the time steps $t_1:t_2$, where
  $1 \le t_1 < t_2 \le T$. Then, for any measurable $f: \mc{W} \to \R$
  \begin{align*}
    \E[f(\outcomes) \mid H_{t_1}(a_{1:t_1-1})]
    = \E\left[ \E_{a_{1:t_2}}^{t_1:t_2} \left[f(\outcomes)\right]
    \mid H_{t_1}(a_{1:t_1-1})\right]
  \end{align*}
  for any $a_{1: t_2} \in \mc{A}_1 \times \cdots \times \mc{A}_{t_2}$.
\end{lemma}
\noindent The following identity---whose proof we give in
Section~\ref{section:proof-conditional-exp}---is a simple consequence of the
definition of conditional expectations, and the tower law.
\begin{lemma}
  \label{lemma:conditional-exp}
  For any measurable function $f: \mc{W} \to \R$, and $1\le t_1 \le t_2\le T$,
  \begin{align*}
    \E_{a_{1:t_2}}^{t_1 : t_2} f(\outcomes)
    = \E\left[f(\outcomes)
    \prod_{t=t_1}^{t_2} \frac{\indic{A_T = a_t}}{\pi_t(a_t \mid H_t(a_{1:t-1}))}
    \mid H_{t_1}(a_{1:t_1-1})
    \right]
  \end{align*}
\end{lemma}

\subsection{Proof of Lemma~\ref{lemma:cand-identity}}
\label{section:proof-cand-identity}

Similar to the notational shorthand~\eqref{eqn:shorthand},
define
\begin{subequations}
\begin{align*}
  \wb{\E}^t_{a_{1:t}}[X]
  \defeq \E[X \mid \bar{H}_t(a_{1:t-1}), \bar{A}_t = a_t] 
  ~~~\mbox{and}~~~
  \wb{\E}^{t:T}_{a_{1:T}}[X]
  = \wb{\E}^{t}_{a_{1:t}}[\wb{\E}^{t+1}_{a_{1:t+1}} [\cdots \wb{\E}^T_{a_{1:T}}[X] \cdots ]].
\end{align*}
\end{subequations}

Begin by noting that by definition of conditional expectation
\begin{align*}
  \E[Y(\bar{A}_{1:T}) \mid \bar{H}_1]
   &  = \sum_{a_1 \in \mc{A}_1} \bar{\pi}(a_1 \mid \bar{H}_1)
    \E[Y(a_1, \bar{A}_{2:T}) \mid \bar{H}_1, \bar{A}_1 = a_1] \\
  & = \sum_{a_1 \in \mc{A}_1} \bar{\pi}(a_1 \mid \bar{H}_1)
    \wb{\E}^1_{a_1}[Y(a_1, \bar{A}_{2:T})], 
\end{align*}
and similarly, conditioning on $\bar{H}_2(a_1) = (S_1, \bar{A}_1 = a_1, S_2(a_1))$
yields
\begin{align*}
  \E[Y(a_1, \bar{A}_{2:T}) \mid \bar{H}_2(a_1)] 
  & = \sum_{a_2 \in \mc{A}_2} \bar{\pi}_2(a_2 \mid \bar{H}_2(a_1))
    \E[Y(a_{1:2}, \bar{A}_{3:T}) \mid \bar{H}_2(a_1), \bar{A}_2 = a_2] \\
  & = \sum_{a_2 \in \mc{A}_2} \bar{\pi}_2(a_2 \mid \bar{H}_2(a_1))
  \wb{\E}^2_{a_{1:2}}[Y(a_{1:2}, \bar{A}_{3:T})].
\end{align*}
From the tower law, the above two equalities yield
\begin{align*}
  \E[Y(\bar{A}_{1:T})] 
  & = \E\Bigg[
    \sum_{a_1 \in\mc{A}_1} \bar{\pi}_1(a_1 \mid \bar{H}_1)
    \E\Bigg[
    \sum_{a_2 \in\mc{A}} \bar{\pi}_2(a_2 \mid \bar{H}_2(a_1)) \cdot
    \E[Y(a_{1:2}, \bar{A}_{3:T}) \mid \bar{H}_2(a_1), \bar{A}_2 = a_2]
    \Bigg\vert \bar{H}_1, \bar{A}_1 = a_1
    \Bigg]
    \Bigg] \\
  & = \E\Bigg[
    \sum_{a_1 \in\mc{A}_1} \bar{\pi}_1(a_1 \mid \bar{H}_1)
    \wb{\E}^1_{a_1}\Bigg[
    \sum_{a_2 \in\mc{A}} \bar{\pi}_2(a_2 \mid \bar{H}_2(a_1))
    \cdot
    \wb{\E}^2_{a_{1:2}}[Y(a_{1:2}, \bar{A}_{3:T})] \Bigg] \Bigg].
\end{align*}
Proceeding iteratively as before and expanding each
$\E[Y(a_{1:t-1}, \bar{A}_{t:T}) \mid \bar{H}_t(a_{1:t-1})]$, we arrive at
\begin{align*}
 & \E[Y(\bar{A}_{1:T})] \\
 &  =  \E\left[ \sum_{a_1 \in \mc{A}_1} \bar{\pi}_1(a_1 \mid \bar{H}_1)
  \wb{\E}^{1}_{a_1}\left[
  \wb{\E}^{2}_{a_{1:2}}\left[
  \sum_{a_2 \in \mc{A}_2} \bar{\pi}_2(a_2 \mid \bar{H}_2(a_1)) \wb{\E}^3_{a_{1:3}}\left[ \cdots
      \sum_{a_T \in \mc{A}_T} \bar{\pi}_T(a_T \mid \bar{H}_T(a_{1:T-1}))
      \wb{\E}^T_{a_{1:T}}\left[
      Y(a_{1:T})
      \right]
  \right]\right]\right]\right].
\end{align*}

Now, we proceed backwards from the inner most expectation to take the outer
sum inside the expectation. By Assumption~\ref{assumption:cand-seq-ig}, we
have
\begin{align*}
  \sum_{a_T \in \mc{A}_T} \bar{\pi}_T(a_T \mid \bar{H}_T(a_{1:T-1}))
  \wb{\E}^T_{a_{1:T}}\left[ Y(a_{1:T}) \right]
  & = \sum_{a_T \in \mc{A}_T} \bar{\pi}_T(a_T \mid \bar{H}_T(a_{1:T-1}))
    \cdot \E\left[ Y(a_{1:T}) ~\Bigg \vert~ \bar{H}_T(a_{1:T-1}) \right] \\
  & = \E\left[ \sum_{a_T \in \mc{A}_T} \bar{\pi}_T(a_T \mid \bar{H}_T(a_{1:T-1}))
    \cdot Y(a_{1:T}) ~\bigg\vert~ \bar{H}_T(a_{1:T-1}) \right].
\end{align*}
Noting that $ \E[ \cdot ~\vert~ \bar{H}_{T}(a_{1:T-1})] = \E[ \cdot ~\vert~ \bar{H}_{T-1}(a_{1:T-2}), S_{T}(a_{1:T-1}), \bar{A}_{T-1}{=}a_{T-1}]$, the tower law and preceding display yield
\begin{align*}
  \wb{\E}^{T-1}_{a_{1:T-1}}
    \left[ \sum_{a_T \in \mc{A}_T} \bar{\pi}_T(a_T \mid \bar{H}_T(a_{1:T-1}))
  \cdot \wb{\E}^T_{a_{1:T}}\left[ Y(a_{1:T}) \right] \right]
  = \wb{\E}^{T-1}_{a_{1:T-1}}
    \left[ \sum_{a_T \in \mc{A}_T} \bar{\pi}_T(a_T \mid \bar{H}_T(a_{1:T-1}))
    \cdot Y(a_{1:T}) \right].
\end{align*}

We repeat an identical process for the sum over $a_{T-1}$. Similarly as above, applying Assumption~\ref{assumption:cand-seq-ig} gives
\begin{align*}
  & \sum_{a_{T-1} \in \mc{A}_{T-1}} \bar{\pi}_{T-1}(a_{T-1} \mid \bar{H}_{T-1}(a_{1:T-2}))
    \cdot \wb{\E}^{T-1}_{a_{1:T-1}}\left[ \sum_{a_T \in \mc{A}_T} \bar{\pi}_T(a_T \mid \bar{H}_T(a_{1:T-1}))
  \cdot Y(a_{1:T}) \right] \\
  & = \E\left[ \sum_{a_{T-1} \in \mc{A}_{T-1}} \bar{\pi}_{T-1}(a_{T-1} \mid \bar{H}_{T-1}(a_{1:T-2}))
    \sum_{a_T \in \mc{A}_T} \bar{\pi}_T(a_T \mid \bar{H}_T(a_{1:T-1}))
    \cdot Y(a_{1:T}) ~\bigg\vert~ \bar{H}_{T-1}(a_{1:T-2}) \right].
\end{align*}
By the tower law, we again get 
\begin{align*}
  & \wb{\E}^{T-2}_{a_{1:T-2}}
    \left[ \sum_{a_{T-1} \in \mc{A}_{T-1}} \bar{\pi}_{T-1}(a_{T-1} \mid \bar{H}_{T-1}(a_{1:T-2}))
    \wb{\E}^{T-1}_{a_{1:T-1}}
    \left[ \sum_{a_T \in \mc{A}_T} \bar{\pi}_T(a_T \mid \bar{H}_T(a_{1:T-1}))
  \cdot Y(a_{1:T}) \right]
   \right] \\
  & = \wb{\E}^{T-2}_{a_{1:T-2}}
    \left[ \sum_{a_{T-1} \in \mc{A}_{T-1}} \bar{\pi}_{T-1}(a_{T-1} \mid \bar{H}_{T-1}(a_{1:T-2}))
    \cdot \sum_{a_T \in \mc{A}_T} \bar{\pi}_T(a_T \mid \bar{H}_T(a_{1:T-1}))
    \cdot Y(a_{1:T}) \right].
\end{align*}

Iterating the above process over the indices $t = T-2, \ldots, 1$, we arrive
at the desired formula.

\subsection{Proof of Lemma~\ref{lemma:seq-ig-nested-exp}}
\label{section:proof-seq-ig-nested-exp}

From the tower law and sequential ignorability of $\pi$,
\begin{align*}
  \E[f(\outcomes) \mid H_{t_1}(a_{1:t_1-1})]
  & = \E[ f(\outcomes) \mid
    H_{t_1}(a_{1:t_1-1}), A_{t_1} = a_{t_1}] \\
  &  =  \E[ \E[f(\outcomes) \mid H_{t_1+1}(a_{1:t_1})] \mid
    H_{t_1}(a_{1:t_1-1}), A_{t_1} = a_{t_1}]
\end{align*}
Applying the tower law to the inner expectation, and applying sequential
ignorability again, we get
\begin{align*}
  \E[f(\outcomes) \mid H_{t_1+1}(a_{1:t_1})]
  = \E\left[\E[ f(\outcomes) \mid
    H_{t_1+2}(a_{1:t_1+1})]
    \mid H_{t_1+1}(a_{1:t_1}), A_{t_1+1} = a_{t_1+1}
  \right]
\end{align*}
Plugging this back into the original display, we have
\begin{align*}
  \E[f(\outcomes) \mid H_{t_1}(a_{1:t_1-1})]
  = \E_{a_1:t_1+1}^{t_1:t_1+1}\left[
  \E[ f(\outcomes) \mid
    H_{t_1+2}(a_{1:t_1+1})]
  \right] 
\end{align*}
Repeating this argument over $t = t_1+2, \ldots, t_2$, we conclude the result.

\subsection{Proof of Lemma~\ref{lemma:conditional-exp}}
\label{section:proof-conditional-exp}

From the definition of conditional expectations, we have
\begin{align*}
  \E[f(\outcomes) \mid H_t(a_{1:t-1}), A_t = a_t]
  = \E\left[
  f(\outcomes) \frac{\indic{A_t = a_t}}{\pi_t(a_t \mid H_t(a_{1:t-1}))}
  \mid H_t(a_{1:t-1})
  \right].
\end{align*}
The result follows by applying this equality at $t = t_2$, applying the tower law,
and iterating the same argument over $t = t_2 - 1, \ldots, t_1$.


\section{Proof of key identities}
\label{section:derivations}

\subsection{Proof of Lemma~\ref{lemma:seq-ig}}
\label{section:proof-seq-ig}

Recalling the notation~\eqref{eqn:averaged-outcome}, sequential ignorability
of $\bar{\pi}$ and Lemma~\ref{lemma:cand-identity} gives the following
representation
\begin{align*}
  \E\left[Y(\bar{A}_{1:T})\right]
    = \sum_{a_{1:T}}  \E\left[ Y(a_{1:T}; \bar{\pi})\right].
\end{align*}
We deal with each term $\E[Y(a_{1:T}; \bar{\pi})]$ in the summation
separately, for each fixed sequence of actions $a_{1:T}$. From sequential
ignorability of $\pi$ and Lemma~\ref{lemma:seq-ig-nested-exp},
\begin{align*}
  \E[Y(a_{1:T}; \bar{\pi})]
  = \E[\E^1_{a_1}[ \cdots \E^T_{a_{1:T}}[ Y(a_{1:T}; \bar{\pi})] \cdots]] 
  = \E[\E^{1:T}_{a_{1:T}}[Y(a_{1:T}; \bar{\pi})]].
\end{align*}
Applying Lemma~\ref{lemma:conditional-exp}, we get
\begin{align*}
  \E[Y(a_{1:T}; \bar{\pi})]
  = \E[ Y(a_{1:T}; \bar{\pi}) \prod_{t=1}^T \frac{\indic{A_t = a_t}}{\pi_t(a_t \mid H_t(a_{1:t-1}))} ].
\end{align*}
Summing the preceeding display over $a_{1:T}$, we obtain the desired result.

\subsection{Proof of Proposition~\ref{prop:single-like-ratio}}
\label{section:proof-single-like-ratio}

From Lemma~\ref{lemma:cand-identity}, we have
\begin{align*}
  \E[Y(\bar{A}_{1:T})]
  = \E\left[
  \sum_{a_{1:T}} Y(a_{1:T}) \prod_{t=1}^T \bar{\pi}_t(a_t \mid \bar{H}_t(a_{1:t-1}))
  \right].
\end{align*}
Since sequential ignorability for $\pi$ holds at any $t < t\opt$,
Lemma~\ref{lemma:seq-ig-nested-exp} implies that the preceeding display is
equal to
\begin{align*}
  \E\left[
  \sum_{a_{1:t\opt-1}} \E_{a_{1:t\opt-1}}^{1:t\opt-1} \left[
  \sum_{a_{t\opt:T}} Y(a_{1:T}) \prod_{t=1}^T \bar{\pi}_t(a_t \mid \bar{H}_t(a_{1:t-1}))
  \right]
  \right].
\end{align*}
Applying Lemma~\ref{lemma:conditional-exp} to the inner expectations, we get
\begin{align*}
  \E[Y(\bar{A}_{1:T})]
  & = \E\left[
  \sum_{a_{1:t\opt-1}} \prod_{t=1}^{t\opt-1} \frac{\indic{A_t = a_t}}{\pi_t(a_t \mid H_t(a_{1:t-1}))}
  \sum_{a_{t\opt:T}} Y(a_{1:T}) \prod_{t=1}^T \bar{\pi}_t(a_t \mid \bar{H}_t(a_{1:t-1}))
    \right] \\
  & = \E\left[
  \prod_{t=1}^{t\opt-1} \frac{\bar{\pi}_t(A_t \mid \bar{H}_t(A_{1:t-1})) }{\pi_t(A_t \mid H_t)}
  \sum_{a_{t\opt:T}} Y(A_{1:t\opt-1}, a_{t\opt:T}) \prod_{t=t\opt}^T \bar{\pi}_t(a_t \mid \bar{H}_t(A_{1:t\opt-1}, a_{t\opt:t-1}))
    \right].
\end{align*}
From the tower law, we arrive at 
\begin{align}
  \E[Y(\bar{A}_{1:T})]
  & = \E\left[ \E\left[
  \prod_{t=1}^{t\opt-1} \frac{\bar{\pi}_t(A_t \mid \bar{H}_t(A_{1:t-1})) }{\pi_t(A_t \mid H_t)}
  \sum_{a_{t\opt:T}} Y(A_{1:t\opt-1}, a_{t\opt:T}) \prod_{t=t\opt}^T \bar{\pi}_t(a_t \mid \bar{H}_t(A_{1:t\opt-1}, a_{t\opt:t-1}))
  ~\Bigg|~ H_{t\opt}
  \right]
  \right] \nonumber \\
  & = \E\left[ \prod_{t=1}^{t\opt-1} \frac{\bar{\pi}_t(A_t \mid \bar{H}_t(A_{1:t-1})) }{\pi_t(A_t \mid H_t)}
    \E\left[
  \sum_{a_{t\opt:T}} Y(A_{1:t\opt-1}, a_{t\opt:T}) \prod_{t=t\opt}^T \bar{\pi}_t(a_t \mid \bar{H}_t(A_{1:t\opt-1}, a_{t\opt:t-1}))
  ~\Bigg|~ H_{t\opt}
  \right]
  \right]. \label{eqn:pre-likelihood}
\end{align}

Applying the  tower law to the inner expectation in the final display, we can write
\begin{align*}
   &  \E\left[
    \sum_{a_{t\opt:T}} Y(A_{1:t\opt-1}, a_{t\opt:T}) \prod_{t=t\opt}^T \bar{\pi}_t(a_t \mid \bar{H}_t(A_{1:t\opt-1}, a_{t\opt:t-1}))
    ~\Bigg|~ H_{t\opt} \right] \\
  & =  \E\left[
    \E\left[
    \sum_{a_{t\opt:T}} Y(A_{1:t\opt-1}, a_{t\opt:T}) \prod_{t=t\opt}^T \bar{\pi}_t(a_t \mid \bar{H}_t(A_{1:t\opt-1}, a_{t\opt:t-1}))
    ~\Bigg|~ H_{t\opt}, A_{t\opt} \right]
    ~\Bigg|~ H_{t\opt} \right] \\
   &  = \sum_{a_{t\opt}, a_{t\opt}'} \bar{\pi}_{t\opt}(a_{t\opt} \mid \bar{H}_{t\opt}(A_{1:t\opt-1}))
     \pi_{t\opt}(a_{t\opt}' \mid H_{t\opt}) \\
   & \qquad \qquad \times \E\left[
    \sum_{a_{t\opt+1:T}} Y(A_{1:t\opt-1}, a_{t\opt:T}) \prod_{t=t\opt+1}^T \bar{\pi}_t(a_t \mid \bar{H}_t(A_{1:t\opt-1}, a_{t\opt:t-1}))
     ~\Bigg|~ H_{t\opt}, A_{t\opt} = a_{t\opt}' \right] \\
  & = \sum_{a_{t\opt}, a_{t\opt}'} \bar{\pi}_{t\opt}(a_{t\opt} \mid \bar{H}_{t\opt}(A_{1:t\opt-1}))
     \pi_{t\opt}(a_{t\opt}' \mid H_{t\opt}) \\
   & \qquad \qquad \times \E\left[ \mc{L}(\outcomes; H_{t\opt}, a_{t\opt}, a_{t\opt}')
    \sum_{a_{t\opt+1:T}} Y(A_{1:t\opt-1}, a_{t\opt:T}) \prod_{t=t\opt+1}^T \bar{\pi}_t(a_t \mid \bar{H}_t(A_{1:t\opt-1}, a_{t\opt:t-1}))
     ~\Bigg|~ H_{t\opt}, A_{t\opt} = a_{t\opt} \right]
\end{align*}
where in the last equality, we used the definition
\begin{equation*}
  \mc{L}(\cdot; H_{t\opt}, a_{t\opt}, a_{t\opt}')
  \defeq \frac{dP_{W}(\cdot \mid H_{t\opt}, A_{t\opt} = a_{t\opt}')}{dP_{W}(\cdot \mid H_{t\opt}, A_{t\opt} = a_{t\opt})}.
\end{equation*}

Again, by the tower law, 
\begin{align*}
  & \E\left[ \mc{L}(\outcomes; H_{t\opt}, a_{t\opt}, a_{t\opt}')
    \sum_{a_{t\opt+1:T}} Y(A_{1:t\opt-1}, a_{t\opt:T}) \prod_{t=t\opt+1}^T \bar{\pi}_t(a_t \mid \bar{H}_t(A_{1:t\opt-1}, a_{t\opt:t-1}))
  ~\Bigg|~ H_{t\opt}, A_{t\opt} = a_{t\opt} \right] \\
  & = \E\Bigg[ \E\Big[ \mc{L}(\outcomes; H_{t\opt}, a_{t\opt}, a_{t\opt}')
    \sum_{a_{t\opt+1:T}} Y(A_{1:t\opt-1}, a_{t\opt:T})  \\
  & \qquad \qquad \qquad \qquad
    \times \prod_{t=t\opt+1}^T \bar{\pi}_t(a_t \mid \bar{H}_t(A_{1:t\opt-1}, a_{t\opt:t-1}))
    \Big| H_{t\opt+1}(A_{1:t\opt-1}, a_{t\opt})\Big]
    ~\Bigg|~ H_{t\opt}, A_{t\opt} = a_{t\opt} \Bigg]
\end{align*}
From sequential ignorability of $\pi$ for $t > t\opt$ and
Lemma~\ref{lemma:seq-ig-nested-exp}, the preceeding display is equal to
\begin{align*}
  \E\left[
    \sum_{a_{t\opt+1:T}}
    \E_{a_{t\opt:T}}^{t\opt+1:T}
    \mc{L}(\outcomes; H_{t\opt}, a_{t\opt}, a_{t\opt}')
    Y(A_{1:t\opt-1}, a_{t\opt:T}) \prod_{t=t\opt+1}^T \bar{\pi}_t(a_t \mid \bar{H}_t(A_{1:t\opt-1}, a_{t\opt:t-1}))
    ~\Bigg|~ H_{t\opt}, A_{t\opt} = a_{t\opt} \right].
\end{align*}
From Lemma~\ref{lemma:conditional-exp}, we can rewrite the above expression as
\begin{align*}
  \E\left[
  \mc{L}(\outcomes; H_{t\opt}, a_{t\opt}, a_{t\opt}')
  Y_{t\opt}(a_{t\opt})
  \prod_{t=t\opt+1}^T \frac{\bar{\pi}_t(A_t \mid \bar{H}_t(A_{1:t\opt-1}, a_{t\opt}, A_{t\opt+1:t}))}{
  \pi_t(A_t \mid H_t(A_{1:t\opt-1}, a_{t\opt}, A_{t\opt+1:t}))}
    ~\Bigg|~ H_{t\opt}, A_{t\opt} = a_{t\opt} \right].
\end{align*}

Plugging these expressions back into the equality~\eqref{eqn:pre-likelihood}, we obtain the result.


\section{Proof of bounds under unobserved confounding}
\label{section:proof-bounds}

\subsection{Naive bound}
\label{section:proof-naive}

We show the below more general result.
\begin{lemma}
  \label{lemma:naive-appx}
  Let Assumptions~\ref{assumption:cand-seq-ig},~\ref{assumption:confounder},~\ref{assumption:bdd-conf} hold.
  Then, we have
  \begin{align*}
    \E[Y(\bar{A}_{1:T})]
    \ge 
    \E\left[ Y(A_{1:T})
    \prod_{t=1}^T \frac{\bar{\pi}_t(A_t \mid \bar{H}_t(A_{1:t-1}))}{\pi_t(A_t \mid H_t) (\Gamma_t^{-1} \indic{Y(A_{1:T}) < 0}
    + \Gamma_t \indic{Y(A_{1:T}) > 0})}
    \right].
  \end{align*}
\end{lemma}
\begin{proof-of-lemma}
From an identical argument as the proof of Lemma~\ref{lemma:seq-ig},
Assumption~\ref{assumption:confounder} yields
\begin{align*}
  \E[Y(\bar{A}_{1:T})] = \E\left[ Y(A_{1:T})
  \prod_{t=1}^T \frac{\bar{\pi}_t(A_t \mid \bar{H}_t(A_{1:t-1}), U_t)}{\pi_t(A_t \mid H_t, U_t)}
  \right].
\end{align*}
From Assumption~\ref{assumption:cand-seq-ig}, the preceeding display is equal to
\begin{align}
  \label{eqn:naive-interim}
  \E[Y(\bar{A}_{1:T})] = \E\left[ Y(A_{1:T})
  \prod_{t=1}^T \frac{\bar{\pi}_t(A_t \mid \bar{H}_t(A_{1:t-1}))}{\pi_t(A_t \mid H_t, U_t)}
  \right].
\end{align}

Now, we bound $\pi_t(A_t \mid H_t, U_t)$ by $\pi_t(A_t \mid H_t)$.
Assumption~\ref{assumption:bdd-conf} implies
\begin{align*}
  \pi_t(a_t \mid H_t, U_t = u_t)
  \pi_t(a_t' \mid H_t, U_t = u_t')
  \le \Gamma_t \pi_t(a_t' \mid H_t, U_t = u_t) \pi_t(a_t \mid H_t, U_t = u_t').
\end{align*}
Multiplying by $p_{U_t}(u_t' \mid H_t)$ on both sides and integrating over $u_t'$, we get
\begin{align*}
  \pi_t(a_t \mid H_t, U_t = u_t)
  \pi_t(a_t' \mid H_t)
  \le \Gamma_t \pi_t(a_t' \mid H_t, U_t = u_t) \pi_t(a_t \mid H_t).
\end{align*}
Summing over $a_t'$ on both sides, we conclude that
\begin{align*}
  \pi_t(a_t \mid H_t, U_t = u_t)
  \le \Gamma_t  \pi_t(a_t \mid H_t).
\end{align*}
almost surely, for any $t, a_t, H_t, u_t$.
Using this relation to lower bound expression~\eqref{eqn:naive-interim}, we obtain the result.
\end{proof-of-lemma}

\subsection{Proof of Theorem~\ref{theorem:loss-min}}
\label{section:proof-loss-min}

By rewriting the original infimization problem over $L(W; H_{t\opt})$ to
$L(W, A_{t\opt+1:T}; H_{t\opt})$, we have 
\begin{align*}
  & \eta\opt(H_{t\opt}; a_{t\opt}) = \\
  & \inf_{L \ge 0}
  \Bigg\{ \E\Bigg[
    L(W, A_{t\opt+1:T}; H_{t\opt})
    Y(A_{1:T})
    \prod_{t=t\opt+1}^T \rho_{t}
    ~\Big|~ H_{t\opt}, A_{t\opt} = a_{t\opt}
    \Bigg]: \E[L(W, A_{t\opt+1:T}; H_{t\opt}) \mid H_{t\opt}, A_{t\opt} = a_{t\opt}] = 1, ~\mbox{and} \\
  & \hspace{20pt} L(w, a_{t\opt+1:T}; H_{t\opt}) = L(w, a_{t\opt+1:T}'; H_{t\opt}),
    \hspace{10pt} L(w, a_{t\opt+1:T}; H_{t\opt}) \le \Gamma L(w', a_{t\opt+1:T}'; H_{t\opt})  
    ~\mbox{a.s. all}~ w, a_{t\opt+1:T}, w', a_{t\opt+1:T}'
    \Bigg\}.
\end{align*}
Relaxing the equality constraint
$L(w, a_{t\opt+1:T}; H_{t\opt}) = L(w, a_{t\opt+1:T}'; H_{t\opt})$, we arrive at
\begin{align*}
  & \eta\opt(H_{t\opt}; a_{t\opt}) \ge \\
  & \inf_{L \ge 0}
  \Bigg\{ \E\Bigg[
    L(W, A_{t\opt+1:T}; H_{t\opt})
    Y(A_{1:T})
    \prod_{t=t\opt+1}^T \rho_{t}
    ~\Big|~ H_{t\opt}, A_{t\opt} = a_{t\opt}
    \Bigg]: \E[L(W, A_{t\opt+1:T}; H_{t\opt}) \mid H_{t\opt}, A_{t\opt} = a_{t\opt}] = 1, ~\mbox{and} \\
  & \hspace{180pt}  L(w, a_{t\opt+1:T}; H_{t\opt}) \le \Gamma L(w', a_{t\opt+1:T}'; H_{t\opt})  
    ~\mbox{a.s. all}~ w, a_{t\opt+1:T}, w', a_{t\opt+1:T}'
    \Bigg\}.
\end{align*}

The preceeding optimization problem is convex, and Slater's condition holds
for $L \equiv 1$.  By strong duality~\citep[Thm.~8.6.1 and Problem
8.7]{Luenberger69}, we obtain the dual formulation
\begin{align*}
  \sup_{\mu} \inf_{L \ge 0}
  \Bigg\{
  \E\Bigg[
    & L(W, A_{t\opt+1:T}; H_{t\opt})
    \left( Y(A_{1:T})
    \prod_{t=t\opt+1}^T \rho_{t} - \mu \right)
    ~\Big|~ H_{t\opt}, A_{t\opt} = a_{t\opt} 
  \Bigg] + \mu: \\
  & \hspace{100pt} L(w, a_{t\opt+1:T}; H_{t\opt}) \le \Gamma L(w', a_{t\opt+1:T}'; H_{t\opt})  
  ~\mbox{a.s. all}~w, a_{t\opt+1:T}, w', a_{t\opt+1:T}'
  \Bigg\}.
\end{align*}
By inspection, the solution to the inner infimum takes the form
\begin{align*}
  L(w, a_{t\opt+1:T}; H_{t\opt})
  = c \left( \Gamma \indic{Y(A_{1:T})
  \prod_{t=t\opt+1}^T \rho_{t} - \mu < 0}
  + \indic{Y(A_{1:T})
  \prod_{t=t\opt+1}^T \rho_{t} - \mu \ge 0}
  \right)
\end{align*}
for some constant $c> 0$.  Let
$\loss_{\Gamma}'(z) \defeq (z)_+ - \Gamma (z)_-$, the derivative of the
weighted squared loss $\loss_{\Gamma}(z) = \half ( \Gamma (z)_-^2 +
(z)_+^2)$. Plugging the preceeding display into the dual formulation, we get
\begin{align*}
  & \sup_{\mu} \inf_{c \ge 0}
  \Bigg\{
  c \E\Bigg[
    \loss_{\Gamma}'\left( Y(A_{1:T})
    \prod_{t=t\opt+1}^T \rho_{t} - \mu \right)
    ~\Big|~ H_{t\opt}, A_{t\opt} = a_{t\opt} 
  \Bigg] + \mu
  \Bigg\} \\
  & = \sup_{\mu} 
  \Bigg\{\mu: \E\Bigg[
    \loss_{\Gamma}'\left( Y(A_{1:T})
    \prod_{t=t\opt+1}^T \rho_{t} - \mu \right)
    ~\Big|~ H_{t\opt}, A_{t\opt} = a_{t\opt} 
  \Bigg] \ge 0
  \Bigg\}.
\end{align*}
Since the function
$\mu \mapsto \E\Bigg[ \loss_{\Gamma}'\left( Y(A_{1:T})
  \prod_{t=t\opt+1}^T \rho_{t} - \mu \right) ~\Big|~
H_{t\opt}, A_{t\opt} = a_{t\opt} \Bigg]$ is strictly decreasing, the optimal
solution (and its value) in the preceeding display is given by the unique zero
of this function.

We now show that the solution to our loss minimization problem
\begin{align*}
  \kappa(H_{t\opt}; a_{t\opt}) = \argmin_{f(H_{t\opt})}
  \Bigg\{ & \E\Bigg[
  \frac{\indic{A_{t\opt} = a_{t\opt}}}{\pi_{t\opt}(a_{t\opt} \mid H_{t\opt})} \times \loss_{\Gamma} \left(
  Y(A_{1:T})
  \prod_{t=t\opt+1}^T \rho_{t} - f(H_{t\opt})
  \right) \Bigg] \\
  & = \E\Bigg[
  \E\left[\loss_{\Gamma} \left(
  Y(A_{1:T})
  \prod_{t=t\opt+1}^T \rho_{t} - f(H_{t\opt})
  \right) ~\Bigg|~ H_{t\opt}, A_{t\opt} = a_{t\opt}\right] \Bigg]
  \Bigg\}
\end{align*}
is in fact the unique zero of the function
$\mu \mapsto \E\Bigg[ \loss_{\Gamma}'\left( Y(A_{1:T})
  \prod_{t=t\opt+1}^T \rho_{t} - \mu \right) ~\Big|~
H_{t\opt}, A_{t\opt} = a_{t\opt} \Bigg]$. The (almost sure) uniqueness of the
solution follows from strong convexity of $\loss_{\Gamma}$. Since the
optimization is over all $H_{t\opt}$-measurable functions, the argmin is given
by
\begin{align*}
  \argmin_{f(H_{t\opt})} 
  ~\E\left[\loss_{\Gamma} \left(
  Y(A_{1:T})
  \prod_{t=t\opt+1}^T \rho_{t} - f(H_{t\opt})
  \right) ~\Bigg|~ H_{t\opt}, A_{t\opt} = a_{t\opt}\right].
\end{align*}
So long as
$\E[ Y(A_{1:T})^2 \prod_{t=t\opt+1}^T \rho_{t}^2
\mid A_{t\opt} = a_{t\opt}, H_{t\opt}] <\infty$ almost surely, first order optimality conditions of
this loss minimization problem is equivalent to
$\E\Bigg[ \loss_{\Gamma}'\left( Y(A_{1:T}) \prod_{t=t\opt+1}^T
  \rho_{t} - f(H_{t\opt}) \right) ~\Big|~ H_{t\opt}, A_{t\opt} =
a_{t\opt} \Bigg] = 0$, which gives our result.

\subsection{Proof of Theorem~\ref{theorem:consistency}}
\label{section:proof-single-consistency}

Our result is based on epi-convergence theory~\cite{KingWe91,
  RockafellarWe98}, which shows (uniform) convergence of convex functions, and
solutions to convex optimization problems.
\begin{definition}
  \label{definition:set-convergence}
  Let $\{A_n\}$ be a sequence of subsets of $\R^d$. The \emph{limit supremum}
  (or \emph{limit exterior} or \emph{outer limit}) and \emph{limit infimum}
  (\emph{limit interior} or \emph{inner limit}) of the sequence $\{A_n\}$
  are
  \begin{align*}
    \limsup_n A_n & \defeq \left\{v \in \R^d
    \mid \liminf_{n \to \infty} \dist(v, A_n) = 0 \right\} ~~~ \mbox{and} \\
    \liminf_n A_n & \defeq \left\{v \in \R^d
    \mid \limsup_{n \to \infty} \dist(v, A_n) = 0 \right\}.
  \end{align*}
\end{definition}
\noindent
The epigraph of a function $h : \R^d \to \R \cup \{+\infty\} $ is
$\epi h \defeq \{(x, t) \in \R^d \times \R \mid h(x) \le t \}$.  We say
$\lim_n A = A_\infty$ if
$\limsup_n A_n = \liminf_n A_n = A_\infty \subset \R^d$. We define a notion of
convergence for functions in terms of their epigraphs.
\begin{definition}
  \label{def:epi-conv}
  A sequence of functions \emph{$h_n$ epi-converges to a function $h$},
  denoted $h_n \cepi h$, if
  \begin{equation}
    \label{eqn:epi-convergence}
    \epi h = \liminf_{n \to \infty} \epi h_n = \limsup_{n \to \infty}
    \epi h_n.
  \end{equation}
\end{definition}
\noindent
If $h$ is proper ($\dom h \neq \varnothing$), epigraphical
convergence~\eqref{eqn:epi-convergence} is characterizaed by pointwise
convergence on a dense set.
\begin{lemma}[Theorem~7.17,~\citet{RockafellarWe98}]
  \label{lemma:epi}
  Let $h_n : \R^d \to \wb{\R}, h : \R^d \to \wb{\R}$ be closed, convex, and
  proper. Then $h_n \cepi h$ is equivalent to either of the following two
  conditions.
  \begin{enumerate}[(i)]
  \item There exists a dense set $A \subset \R^d$ such that
    $h_n(v) \to h(v)$ for all $v \in A$.
  \item For all compact $C \subset \dom h$ not containing a boundary point
    of $\dom h$,
    \begin{equation*}
      \lim_{n \to \infty} \sup_{v \in C} |h_n(v) - h(v)| = 0.
    \end{equation*}
  \end{enumerate}
\end{lemma}
\noindent The last characterization of epigraphical convergence is powerful as
it gives convergence of solution sets.
\begin{lemma}[Theorem 7.31, \citet{RockafellarWe98}]
  \label{lemma:epi-solution-consistency}
  Let $h_n : \R^d \to \wb{\R}, h : \R^d \to \wb{\R}$ satisfy $h_n \cepi h$
  and $-\infty < \inf h < \infty$. Let $S_n(\varepsilon) = \{\theta \mid
  h_n(\theta) \le \inf h_n + \varepsilon\}$ and $S(\varepsilon) = \{\theta
  \mid h(\theta) \le \inf h + \varepsilon\}$.  Then $\limsup_n
  S_n(\varepsilon) \subset S(\varepsilon)$ for all $\varepsilon \ge 0$, and
  $\limsup_n S_n(\varepsilon_n) \subset S(0)$ whenever $\varepsilon_n
  \downarrow 0$.
\end{lemma}

From Lemmas~\ref{lemma:epi},~\ref{lemma:epi-solution-consistency}, it suffices
to show that the expected loss function and its empirical counterpart
satisfies appropriate regularity conditions (proper and closed), and show that
our empirical loss pointwise converges to the population loss almost
surely. Recall that $\mc{D}_n$ is the split of data used to estimate $\what{\pi}$, and let $\mc{D}_{\infty}$ be the $\sigma$-algebra defined by $\mc{D}_n$ as
$n \to \infty$.  Our subsequent argument will be conditional on
$\mc{D}_{\infty}$, and the event
\begin{equation*}
  \event \defeq \left\{
    \what{\pi}_t \to \pi_t~~\mbox{pointwise},~ \what{\rho}_{t} \le 2C ,~\mbox{and}~~
    \what{\pi}_{t\opt}(a_{t\opt} \mid H_{t\opt}) \in [(2C)^{-1}, 1]
  \right\}.
\end{equation*}
We assume w.l.o.g. (increasing $C$ if necessary) that $c \le (2C)^{-1}$. Note that $\P(\event) = 1$ by assumption.

First, note that since $\theta \mapsto f_{\theta}$ is
linear,
$\theta \mapsto \loss_{\Gamma}( Y(A_{1:T}) \prod_{t=t\opt+1}^T
\what{\rho}_{t} - f_{\theta}(H_{t\opt}))$ is convex.  Both the empirical and
population loss
\begin{align*}
  & \theta \mapsto \what{\E}_n\left[
  \frac{\indic{A_{t\opt} = a_{t\opt}}}{\what{\pi}_{t\opt}(a_{t\opt} \mid H_{t\opt})}
  \loss_{\Gamma} \left(
  Y(A_{1:T})
  \prod_{t=t\opt+1}^T \what{\rho}_{t} - f_{\theta}(H_{t\opt})
    \right) \right] \eqdef \what{h}_n(\theta), \\
  & \theta \mapsto \E\left[
    \frac{\indic{A_{t\opt} = a_{t\opt}}}{\pi_{t\opt}(a_{t\opt} \mid H_{t\opt})}
    \loss_{\Gamma} \left(
    Y(A_{1:T})
    \prod_{t=t\opt+1}^T \rho_{t} - f_{\theta}(H_{t\opt})
    \right) \right] \eqdef h(\theta),
\end{align*}
are proper since they are nonnegative, and finite a.s. at $\theta = 0$.
Since the functions
\begin{align*}
  & \theta \mapsto 
  \frac{\indic{A_{t\opt} = a_{t\opt}}}{\what{\pi}_{t\opt}(a_{t\opt} \mid H_{t\opt})}
  \loss_{\Gamma} \left(
  Y(A_{1:T})
  \prod_{t=t\opt+1}^T \what{\rho}_{t} - f_{\theta}(H_{t\opt})
    \right), \\
  & \theta \mapsto 
    \frac{\indic{A_{t\opt} = a_{t\opt}}}{\pi_{t\opt}(a_{t\opt} \mid H_{t\opt})}
    \loss_{\Gamma} \left(
    Y(A_{1:T})
    \prod_{t=t\opt+1}^T \rho_{t} - f_{\theta}(H_{t\opt})
    \right),
\end{align*}
are continuous by linearity of $\theta \mapsto f_{\theta}$, dominated
convergence shows continuity of both the empirical loss $\what{h}_n(\theta)$ (a.s.) and population loss $h(\theta)$.

Next, we show that the empirical plug-in loss converges pointwise to its
population counterpart almost surely. Since $S(0) = \{\theta\opt\}$ by
hypothesis, Lemmas~\ref{lemma:epi},~\ref{lemma:epi-solution-consistency} will
give the final result. Defining the function
\begin{align*}
  h_n(\theta) \defeq 
  \E\left[ \frac{\indic{A_{t\opt} = a_{t\opt}}}{\what{\pi}_{t\opt}(a_{t\opt} \mid H_{t\opt})}
  \loss_{\Gamma} \left(
  Y(A_{1:T})
  \prod_{t=t\opt+1}^T \what{\rho}_{t} - f_{\theta}(H_{t\opt})
  \right) \right],
\end{align*}
we write
\begin{equation*}
  |\what{h}_n(\theta) - h(\theta)|
  \le |h_n(\theta) -  h(\theta)| + |\what{h}_n(\theta) -  h_n(\theta)|,
\end{equation*}
and show that each term in the right hand side converges to $0$ almost
surely.

To show that the first term goes to zero, note that since $\what{\pi}_{t\opt} \to \pi_{t\opt}$ a.s., we have $\pi_{t\opt}(a_{t\opt} \mid H_t) \ge (2C)^{-1}$ a.s. for all $a_{t\opt}$. This gives
\begin{align*}
  |h_n(\theta) -  h(\theta)| 
 & \le \left|h_n(\theta) - \E\left[\frac{\indic{A_{t\opt} = a_{t\opt}}}{\pi_{t\opt}(a_{t\opt} \mid H_{t\opt})}
  \loss_{\Gamma} \left(
  Y(A_{1:T})
  \prod_{t=t\opt+1}^T \what{\rho}_{t} - f_{\theta}(H_{t\opt})
  \right)\right]\right| \\
  & \qquad + \left| \E\left[\frac{\indic{A_{t\opt} = a_{t\opt}}}{\pi_{t\opt}(a_{t\opt} \mid H_{t\opt})}
  \loss_{\Gamma} \left(
  Y(A_{1:T})
  \prod_{t=t\opt+1}^T \what{\rho}_{t} - f_{\theta}(H_{t\opt})
  \right)\right] - h(\theta)\right| \\
  & \le \Gamma C^2 \E\Bigg[ \left|\pi_{t\opt}(a_{t\opt} \mid H_{t\opt}) - \what{\pi}_{t\opt}(a_{t\opt} \mid H_{t\opt})\right|
    \cdot \left( Y(A_{1:T})^2 (2C)^{2T} + 2 |f_{\theta}(H_{t\opt})|^2  \right) \\
  & \qquad + 
  \Gamma C \E\Bigg[ \left( Y(A_{1:T}) 2(2C)^{T} + 2 |f_{\theta}(H_{t\opt})| 
  \right)
  \cdot Y(A_{1:T})  \cdot \left|
  \prod_{t=t\opt+1}^T \what{\rho}_{t}
   - \prod_{t=t\opt+1}^T \rho_{t} 
  \right|
  \Bigg],
\end{align*}
which has an integrable envelope function under our assumptions and conditional on $\event$.
By dominated convergence, we have the result since $\what{\pi_t} \to \pi_t$
almost surely (and hence $\what{\rho}_{t} \cas \rho_{t}$).

To show that the second term converges to zero, we use the following strong law
of large numbers for triangular arrays.
\begin{lemma}[{\citet[Theorem 2]{HuMoTa89}}]
  \label{lemma:triangular-slln}
  Let $\{\xi_{ni}\}_{i=1}^n$ be a triangular array where
  $X_{n1}, X_{n2}, \ldots$ are independent random variables for any fixed
  $n$. If there exists $\xi$ such that $|\xi_{ni}|\le\xi$ and
  $\E[\xi^2] <\infty$, then
  $\frac{1}{n} \sum_{i=1}^n (\xi_{ni} - \E[\xi_{ni}]) \cas 0$.
\end{lemma}
The random variable 
\begin{align*}
  \frac{\indic{A_{t\opt} = a_{t\opt}}}{\what{\pi}_{t\opt}(a_{t\opt} \mid H_{t\opt})}
  \loss_{\Gamma} \left(
  Y(A_{1:T})
  \prod_{t=t\opt+1}^T \what{\rho}_{t} - f_{\theta}(H_{t\opt})
  \right)
\end{align*}
are i.i.d. for each trajectory, conditional on $\mc{D}_{\infty}$. By
convexity, the below random variable upper bounds the preceeding display
\begin{equation*}
  \xi = 16\Gamma (2C)^{2T}
  \left(  f_{\theta}(H_{t\opt})^2 +
    Y(A_{1:T})^2 
  \right)
\end{equation*}
on the event $\event$. From hypothesis, we have
$\E[\xi^2 \mid \mc{D}_{\infty}, \event] < \infty$.  Applying
Lemma~\ref{lemma:triangular-slln} conditional on $\mc{D}_{\infty}$ and $\event$,
we conclude
\begin{align*}
  |\what{h}_n(\theta) -  h_n(\theta)| \cas 0.
\end{align*}

Applying Lemmas~\ref{lemma:epi},~\ref{lemma:epi-solution-consistency}, we
conclude that for any $\varepsilon_n \downarrow 0$,
$\liminf_{n \to \infty} \mbox{dist}(\theta\opt, S_{\varepsilon_n}) \cp 0$
conditional on $\mc{D}_{\infty}$ and $\event$.
Now, note that for any $\Delta > 0$,
\begin{align*}
  \P\left( |\liminf_{n \to \infty} \mbox{dist}(\theta\opt, S_{\varepsilon_n})| \ge \Delta \right)
  & = \P\left( |\liminf_{n \to \infty} \mbox{dist}(\theta\opt, S_{\varepsilon_n})|
    \ge \Delta \mid \event \right)  \\
  & = \E\left[ \P\left( |\liminf_{n \to \infty} \mbox{dist}(\theta\opt, S_{\varepsilon_n})|
    \ge \Delta \mid \mc{D}_{\infty}, \event \right) \mid \event \right] \\
  & = \E\left[ \P\left( |\liminf_{n \to \infty} \mbox{dist}(\theta\opt, S_{\varepsilon_n})|
    \ge \Delta \mid \mc{D}_{\infty}, \event \right) \right]
\end{align*}
where the first and the last equality used since $\P(\event) = 1$. By dominated
convergence, the preceeding display goes to $0$ as $n \to \infty$.


\section{Experiments}
This section provides implementation details for the experiments presented in the main text.

\begin{figure}[!tb]
  \centering  
  \begin{tabular}{cc}
    \includegraphics[width=.45\linewidth]{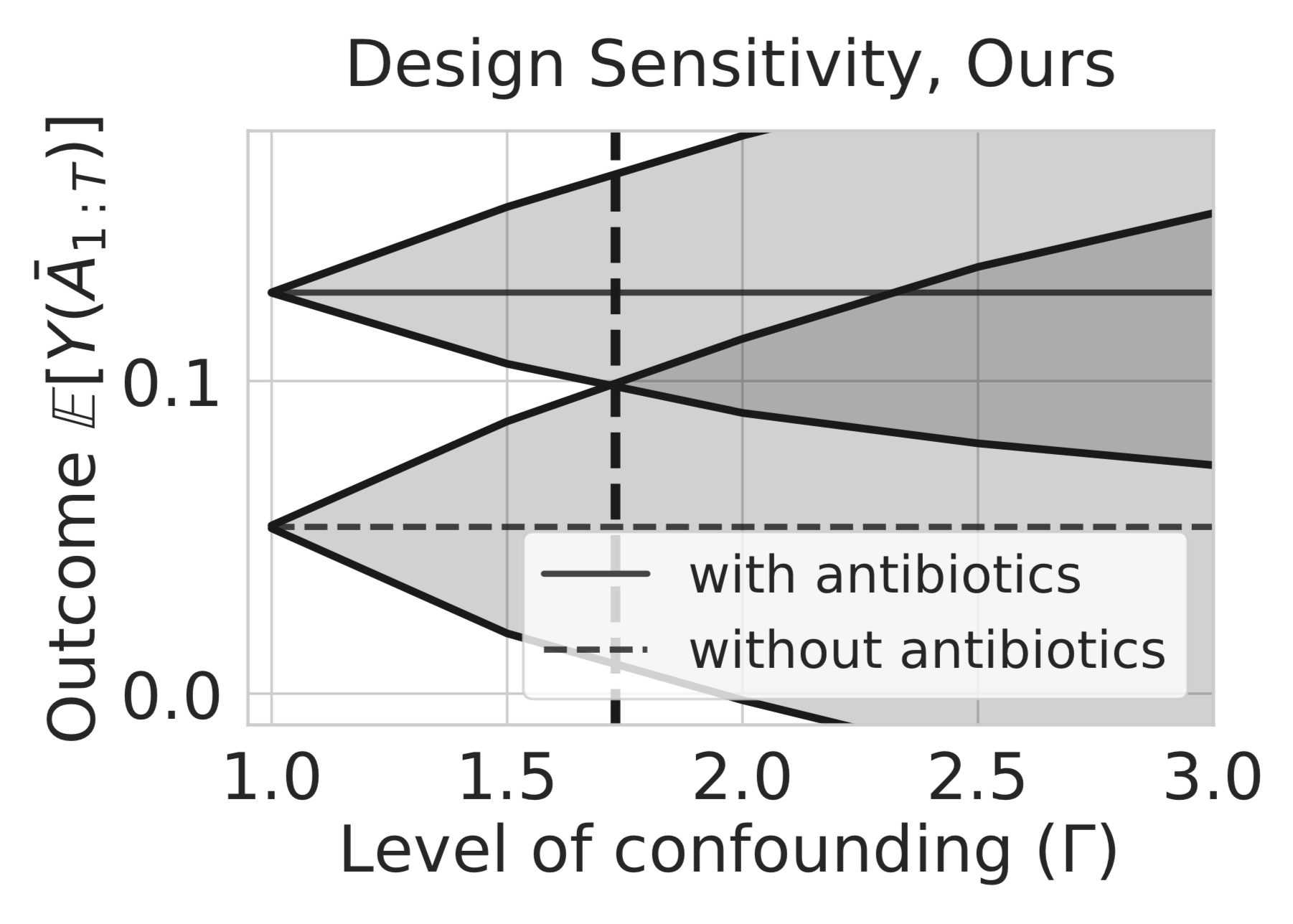}      
    &     
    \includegraphics[width=.45\linewidth]{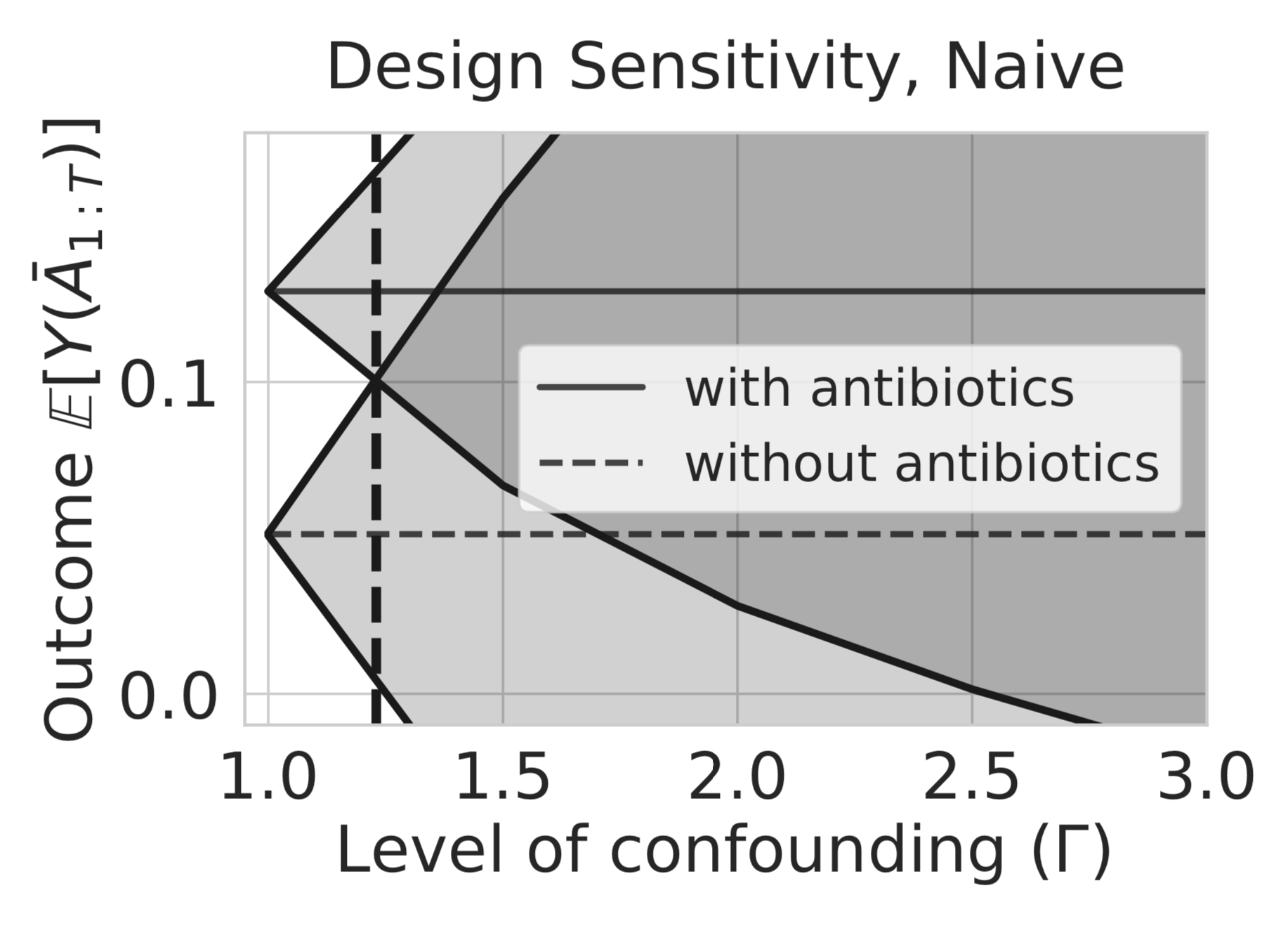} \\  
    {\small (a) Our approach} & {\small (b) Naive approach}
  \end{tabular}                                                  
  \caption{\label{fig:appx:sepsis_design}Sepsis simulator design sensitivity. Data generation process with level of confounding $\Gamma^\star = 1.0$. Estimated lower and upper bound of two policies (with and without antibiotics) under (a) our approach with sensitivity $1.7$ (b) naive approach with sensitivity $1.23$.}       
  \vspace{-4 mm}
\end{figure}    

\subsection{Sepsis Sim}
We use the sepsis simulator developed by~\citet{OberstSo2019}.

\paragraph{The optimal policy} Recall that we assume that the decisions are
made near-optimally. To learn the optimal policy, we generate 2000 samples for
each transition and constructed the transition matrix $P(s,a,s')$ and the
reward matrix $R(s,a,s')$ of the MDP. Similar to \citet{OberstSo2019} we used
policy iteration to learn the optimal policy. We create a near-optimal (soft
optimal) policy by having the policy take a random action with probability
$0.05$, and the optimal action with probability $0.95$. The value function
(for the optimal policy) was computed using value iteration. The horizon is
$T=5$ and the discount factor $\gamma = 0.99$, which results in soft optimal
policy having an average value (over the possible distribution of state
states) of $0.14$.

\paragraph{Confounding} We injected confounding in the first decision of this
simulation by defining two different policies: ``with antibiotics" and
``without antibiotics". ``with antibiotics" which is identical to the soft
optimal policy except that the probability mass of actions without antibiotics
is moved to the corresponding action with antibiotics. For example, if the
probability of the action $a_1= $(antibiotics on, vasopressors off,
ventilation on) in the soft optimal policy is $p_1$, and $a_1' = $(antibiotics
off, vasopressors off, ventilation on) is $p_1'$, then in the ``with
antibiotics" $a_1$ has probability $p_1 + p_1'$ and $a_1'$ has probability
zero in this new policy. The ``without antibiotics" does the opposite: moves
probability mass of actions with antibiotics to the corresponding action
without antibiotics. In our confounding scenario, for healthy patients we administer antibiotics (i.e. follow the ``with antibiotics'') policy with a higher
probability (w.p. $\frac{\sqrt{\Gamma}}{1+\sqrt{\Gamma}}$). For unhealthy patients, we administer antibiotics with a lower probability (w.p. $\frac{1}{1+\sqrt{\Gamma}}$).

Concretely, to compute the transition from a state conditional on an action,
we do inverse transform sampling: we generate a uniform random variable $U_t$
on $[0,1]$, and use this to index into the transition probability distribution
for the next state, sorted by the states' value function and current reward.
This coupling ensures that if $U_t$ is large, then the next state will have a
high value, and if $U_t$ is small, then the next state will have a low
value. The hidden variable $U$ used for confounding in the first decision is
$U = \sum_{t=1}^T U_t$, which serves as a surrogate for the health of patient,
because the larger $U$ is, the more likely the patient is to have improving
state values. We choose a threshold $u_0$, and if $U > u_0$, the behavior
policy follows the action with antibiotics, and if $U \le u_0$, the behavior
policy follows the action without antibiotics, thus introducing confounding.

After the first decision, the behaviour policy is a mixture of two policies:
$85\%$ the soft optimal policy and $15\%$ of a sub-optimal policy that is
similar to the soft optimal but the vasopressors action is flipped. For
example, if probability of the action $a_1= $(antibiotics on, vasopressors
off, ventilation on) is $p_1$, and $a_1'= $(antibiotics on, vasopressors on,
ventilation on) in $p_1'$ in the soft optimal policy, then the sub-optimal has
probability $p_1'$ and $p_1$ for action $a_1$ and $a_1'$, respectively.

\paragraph{Loss minimization} Since the state and action space are discrete,
we learn the tabular value $\kappa(s, a)$ for each state action pair
separately to minimize the empirical loss. Additionally, in order to compute
the upper bound of both ours and the naive method, we compute the negative of
the lower bound on the negative of return (cost).

\paragraph{Behaviour Policy} We estimate the behaviour policies from the data
in two parts: the first time step and time steps $t=2$ through $t=5$. By the
assumptions stated above, each of these policies depends only on the previous
state, and we learn the tabular probability of each state action pair
$\pi_t(a|s)$ separately.

\paragraph{Design Sensitivity} We present another design sensitivity
experiment, with $\Gamma^{\opt} = 1.0$. Figure~\ref{fig:appx:sepsis_design}
(a,b) shows design sensitivity of our method ($1.7$) versus the na\"ive method
($1.23$).

\subsection{Autism}
\label{app:autism}
In the autism experiments, our data generation process (simulator) is adopted
from \citet[Appendix B]{LuNaKaOsPeFaAl2016}.  Each individual has a set of
covariates $X$, consisting of six mean-centered features: \{\texttt{age},
\texttt{gender}, \texttt{indicator of African American}, \texttt{indicator of
  Caucasian}, \texttt{indicator of Hispanic}, \texttt{indicator of
  Asian}\}. The Autism SMART trial \cite{KasariKaGoNiMaLaMuAl2014} simulator
specifies a set of 300 individuals: to obtain a sample size $N$, we sample
with replacement from this set. For details on the simulator, we refer to
Appendix B of \citet{LuNaKaOsPeFaAl2016}. At the first timestep there are two
actions available $A_1 \in \{-1, 1\}$, where $A_1 = 1$ denote BLI, and $A_1=-1$ denote AAC. At the second timestep there are
three actions $A_2 \in \{-1, 0, 1\}$, where $A_2=1$ denote assigning intensified BLI to slow responders, $A_2=-1$ denote assigning AAC to slow responder and $A_2=0$ denote continuing with the same action for fast responders.

\paragraph{Confounding} The original simulator did not have confounding. We
now describe how we introduce confounding in this setting.

\citet[Appendix B]{LuNaKaOsPeFaAl2016} specifies the effect of the second
action (whether to augment BLI with AAC) on the reward outcome $Y$ as follows:
\begin{align*}
    Y = \eta^T_{31}X + \eta_{22}Y_0 + \eta_{33}^T A_1 + \eta_{34}Y_{12} - 2 \theta (1-R)(A_1+1)A_2 + \epsilon. 
\end{align*}
$A_1$ is either $-1$ or $1$. Therefore the final term (outside of the noise
$\epsilon$) is non-zero only when $A_1=1$, and we can interpret $\theta$ as
the effect size of the adaptive policy (which always takes $A_1=1$); for exact
definition of the effect size refer to \citet{LuNaKaOsPeFaAl2016}. For those
more familiar with the RL literature, it is related to the advantage
function. In the original paper, Figure 7 in \citet{LuNaKaOsPeFaAl2016} were
generated using 4 different values of $\theta$.

We introduce confounding by varying $\theta$ (thereby impacting the potential
outcome) and then altering the behavioral treatment decisions according to the
knowledge of that $\theta$. More precisely, given a $\theta_0$, for each
individual, we randomly set $\theta_0 + \sigma_{\theta}$ or
$\theta_0 - \sigma_{\theta}$. The second action is 1 with probability
$\frac{\sqrt{\Gamma}}{1+\sqrt{\Gamma}}$ if $\theta \geq \theta_0$ and 1 with
probability $\frac{1}{1+\sqrt{\Gamma}}$ if $\theta \leq \theta_0$. In our
experiments, we take $\sigma_{\theta} = 5$.

\paragraph{Loss minimization} To estimate $\kappa(H_{t\opt}; a_{t\opt})$ in
the loss minimization problem, we used a neural network with 3 hidden layers
of size \{\texttt{128, 128, 128, 64}\} with \texttt{Relu} activations,
followed by a single linear output layer. We initialize the layers with Xavier
initialization and used the Adam optimizer with learning rate $10^{-3}$. The
input $H_t$ is 10-dimensional consisting of 6 covariates, indicator of slow
responder, initial action $A_1$, number of speech utterances after the initial
action, and an interaction term between $A_1$ and the slow responder
indicator.

\paragraph{Behaviour Policy} We use logistic regression to estimate the
behaviour policy from the observed data: note that this is not the true
behavior policy, because that depends on the (latent) confounding. Different
models were fit for the first and second time steps. For the first timestep
the learned model is $\pi_1(A_1|H_1)$, where $H_1$ contains the observed $X$
(6 covariates), and $A_1 \in \{-1, 1\}$. For the estimated behavior policy in
the second timestep $\pi_2(A_2|H_2)$, $H_2$ includes $X$ (6 covariates), the
action $A_1$, indicator of slow responder, the interaction term between $A_1$
and the indicator, and the number of speech utterances after the initial
action.




\end{document}